\documentclass[acmsmall,screen]{acmart}

\AtBeginDocument{%
  \providecommand\BibTeX{{%
    \normalfont B\kern-0.5em{\scshape i\kern-0.25em b}\kern-0.8em\TeX}}}

\setcopyright{acmcopyright}
\acmJournal{CSUR}
\acmYear{2020} \acmVolume{1} \acmNumber{1} \acmArticle{1} \acmMonth{1} \acmPrice{15.00}\acmDOI{10.1145/3439950}

\usepackage{amsmath}
\usepackage{booktabs}
\usepackage{amsthm}
\usepackage{adjustbox}
\usepackage[font=small]{caption}
\usepackage{multirow} % For formal tables
\usepackage{algorithmic}
\usepackage{algorithm}
\usepackage{graphicx}  

\usepackage{url}
\setcitestyle{numbers,sort&compress}

\newcommand{\eg}{\textit{e.g.}}
\newcommand{\ie}{\textit{i.e.}}
\newcommand{\xvec}{{\mathbf{x}}}
\newcommand{\zvec}{{\mathbf{z}}}
\newcommand{\ysubx}{{y_\mathbf{x}}}
\DeclareMathOperator*{\argmin}{arg\,min}
\DeclareMathOperator*{\argmax}{arg\,max}

\begin{document}

% \title{A Survey on Deep Learning for Anomaly Detection}
%
% CS：
\title{Deep Learning for Anomaly Detection: A Review}

\author{Guansong Pang}
\authornote{Corresponding author: Guansong Pang, pangguansong@gmail.com}
% \email{guansong.pang@adelaide.edu.au}
\affiliation{%
  \institution{University of Adelaide}
  \city{Adelaide}
  \state{South Australia}
  \postcode{5005}
}

\author{Chunhua Shen}
% \email{chunhua.shen@adelaide.edu.au}
\affiliation{%
  \institution{University of Adelaide}
  \city{Adelaide}
  \state{South Australia}
  \postcode{5005}
}

\author{Longbing Cao}
% \email{longbing.cao@uts.edu.au}
\affiliation{%
  \institution{University of Technology Sydney}
  \city{Sydney}
  \state{New South Wales}
  \postcode{2007}
}

\author{Anton van den Hengel}
% \email{anton.vandenhengel@adelaide.edu.au}
\affiliation{%
  \institution{University of Adelaide}
  \city{Adelaide}
  \state{South Australia}
  \postcode{5005}
}
\renewcommand{\shortauthors}{Pang, et al.}

% \author{Guansong Pang

% \thanks{Guansong Pang is with The University of Adelaide, Australia.}
\begin{abstract}
Anomaly detection, a.k.a. outlier detection or novelty detection, has been a lasting yet active research area in various research communities for several decades. There are still some unique problem complexities and challenges that require advanced approaches. In recent years, deep learning enabled anomaly detection, \ie, \textit{deep anomaly detection}, has emerged as a critical direction. This paper surveys the research of deep anomaly detection with a comprehensive taxonomy, covering advancements in three high-level categories and 11 fine-grained categories of the methods. We review their key intuitions, objective functions, underlying assumptions, advantages and disadvantages, and discuss how they address the aforementioned challenges. We further discuss a set of possible future opportunities and new perspectives on addressing the challenges.

% In recent years deep-learning-enabled anomaly detection has shown tremendous improvement over traditional methods, drawing significant attention in diverse research communities, but there is a lack of systematic review and discussion of the research progress in this area. In this work we aim to present a comprehensive review of this area to unravel existing methods and discuss future opportunities. Particularly, we introduce a taxonomy of deep anomaly detection that formulate and categorize a large number of state-of-the-art methods into three high-level categories and 11 fine-grained categories. We then review the methodology, assumption, technical details, advantages and disadvantages of the methods to enable a comprehensive understanding of the research progress. We further discuss a set of largely unsolved challenging problems and possible future opportunities. We also solicit a collection of publicly accessible source codes of nearly all categories of the methods and a large number of real-world datasets with real anomalies to offer some empirical comparison benchmarks for this area.
\end{abstract}

% \keywords{datasets, neural networks, gaze detection, text tagging}
\maketitle


% \begin{IEEEkeywords}
% Communications Society, IEEE, IEEEtran, journal, \LaTeX, paper, template.
% \end{IEEEkeywords}

% \IEEEpeerreviewmaketitle

\section{Introduction}

Anomaly detection, a.k.a. outlier detection or novelty detection, is referred to as the process of detecting data instances that significantly deviate from the majority of data instances. Anomaly detection has been an active research area for several decades, with early exploration dating back as far as to 1960s \cite{grubbs1969outlier}. Due to the increasing demand and applications in broad domains, such as risk management, compliance, security, financial surveillance, health and medical risk, and AI safety, anomaly detection plays increasingly important roles, highlighted in various communities including data mining, machine learning, computer vision and statistics. In recent years, deep learning has shown tremendous capabilities in learning expressive representations of complex data such as high-dimensional data, temporal data, spatial data and graph data, pushing the boundaries of different learning tasks. Deep learning for anomaly detection, \textit{deep anomaly detection} for short, aim at learning feature representations or anomaly scores via neural networks for the sake of anomaly detection. A large number of deep anomaly detection methods have been introduced, demonstrating significantly better performance than conventional anomaly detection on addressing challenging detection problems in a variety of real-world applications. This work aims to provide a comprehensive review of this area. We first discuss the problem nature of anomaly detection and major largely unsolved challenges, then systematically review the current deep methods and their capabilities in addressing these challenges, and finally presents a number of future opportunities. 

As a popular area, a number of studies \cite{hodge2004survey,chandola2009anomaly,aggarwal2017outlieranalysis,zimek2012survey,akoglu2015graphsurvey,gupta2013temporaloutliersurvey,boukerche2020outlier} have been dedicated to the categorization and review of anomaly detection techniques. However, they all focus on conventional anomaly detection methods only. One work closely related to ours is \cite{chalapathy2019survey}. It presents a good summary of a number of real-world applications of deep anomaly detection, but only provides some very high-level outlines of selective categories of the techniques, from which it is difficult, if not impossible, to gain the sense of the approaches taken by the current methods and their underlying intuitions. By contrast, this review delineates the formulation of current deep detection methods to gain key insights about their intuitions, inherent capabilities and weakness on addressing some largely unsolved challenges in anomaly detection. This forms a deep understanding of the problem nature and the state-of-the-art, and brings about genuine open opportunities. It also helps explain why we need deep anomaly detection.
% Particularly, we systematically formulate and categorize a large number of state-of-the-art methods in the area and present extensive discussions of the key intuitions and optimization objectives of these methods. We further provide important insights into the challenges these methods address and discuss tremendous future opportunities in the area. 

In summary, this work makes the following five major contributions:
\begin{itemize}
    \item \textit{Problem nature and challenges}. We discuss some unique problem complexities underlying anomaly detection and the resulting largely unsolved challenges.
    \item \textit{Categorization and formulation}. We formulate the current deep anomaly detection methods into three principled frameworks: deep learning for generic feature extraction, learning representations of normality, and end-to-end anomaly score learning. A hierarchical taxonomy is presented to categorize the methods based on 11 different modeling perspectives.
    \item \textit{Comprehensive literature review}. We review a large number of relevant studies in leading conferences and journals of several relevant communities, including machine learning, data mining, computer vision and artificial intelligence, to present a comprehensive literature review of the research progress. To provide an in-depth introduction, we delineate the basic assumptions, objective functions, key intuitions and their capabilities in addressing some of the aforementioned challenges by all categories of the methods.
    \item \textit{Future opportunities}. We further discuss a set of possible future opportunities and their implication to addressing relevant challenges.
    \item  \textit{Source codes and datasets}. We solicit a collection of publicly accessible source codes of nearly all categories of methods and a large number of real-world datasets with \textit{real anomalies} to offer some empirical comparison benchmarks.
\end{itemize}

\section{Anomaly Detection: Problem Complexities and Challenges}

Owing to the unique nature, anomaly detection presents distinct problem complexities from the majority of analytical and learning problems and tasks. This section summarizes such intrinsic complexities and unsolved detection challenges in complex anomaly data.

\subsection{Major Problem Complexities}\label{subsec:complexity}

Unlike those problems and tasks on majority, regular or evident patterns, anomaly detection addresses minority, unpredictable/uncertain and rare events, leading to some unique problem complexities to all (both deep and shallow) detection methods: 
% render general deep learning techniques ineffective.

\begin{itemize}
    \item \textbf{Unknownness}. Anomalies are associated with many unknowns, e.g., instances with unknown abrupt behaviors, data structures, and distributions. They remain unknown until actually occur, such as novel terrorist attacks, frauds and network intrusions. 
    % As a result, it is often impossible to gain their supervisory information in advance.
    % ., but training powerful deep learning models typically requires a large amount of manually labeled data.
    \item \textbf{Heterogeneous anomaly classes}. Anomalies are irregular, and thus, one class of anomalies may demonstrate completely different abnormal characteristics from another class of anomalies. For example, in video surveillance, the abnormal events robbery, traffic accidents and burglary are visually highly different.
    \item \textbf{Rarity and class imbalance}. Anomalies are typically rare data instances, contrasting to normal instances that often account for an overwhelming proportion of the data. Therefore, it is difficult, if not impossible, to collect a large amount of labeled abnormal instances. This results in the unavailability of large-scale labeled data in most applications. The class imbalance is also due to the fact that misclassification of anomalies is normally much more costly than that of normal instances.  % to enable the learning of normality/abnormality representations.
    % \item \textbf{Diverse anomaly definition}. Characteristics such as rarity and irregularity are widely-accepted general terms used to define anomalies, but there are a number of heuristics used to quantify these characteristics, resulting in diverse definitions of anomaly in different methods, \eg, distance-, density-, clustering- and isolation-based methods \cite{aggarwal2017outlieranalysis}. Due to this difference, the results of an anomaly detection method can be very different from the other methods.
    \item \textbf{Diverse types of anomaly}. Three completely different types of anomaly have been explored \cite{chandola2009anomaly}. \textit{Point anomalies} are individual instances that are anomalous w.r.t. the majority of other individual instances, e.g., the abnormal health indicators of a patient. \textit{Conditional anomalies}, a.k.a. contextual anomalies, also refer to individual anomalous instances but in a specific context, \ie, data instances are anomalous in the specific context, otherwise normal. The contexts can be highly different in real-world applications, \eg, sudden temperature drop/increase in a particular temporal context, or rapid credit card transactions in unusual spatial contexts. \textit{Group anomalies}, a.k.a. collective anomalies, are a subset of data instances anomalous as a whole w.r.t. the other data instances; the individual members of the collective anomaly may not be anomalies, \eg, exceptionally dense subgraphs formed by fake accounts in social network are anomalies as a collection, but the individual nodes in those subgraphs can be as normal as real accounts.
    % \item \textbf{Costly anomaly-supervisory data collection}. Collecting labeled abnormal instances can be highly costly in many real-world applications,% such as intrusion detection, fraud detection and crime detection, 
    % it is even more costly or impossible to obtain large-scale labeled anomaly data and novel anomalies.
\end{itemize}

\subsection{Main Challenges Tackled by Deep Anomaly Detection}

The above complex problem nature leads to a number of detection challenges. Some challenges, such as scalability w.r.t. data size, have been well addressed in recent years, while the following are largely unsolved, to which deep anomaly detection can play some essential roles.

\begin{itemize}
    \item \textbf{CH1: Low anomaly detection recall rate}. Since anomalies are highly rare and heterogeneous, it is difficult to identify all of the anomalies. Many normal instances are wrongly reported as anomalies while true yet sophisticated anomalies are missed. Although a plethora of anomaly detection methods have been introduced over the years, the current state-of-the-art methods, especially unsupervised methods (\eg, \cite{breunig2000lof,liu2012iforest}), still often incur high false positives on real-world datasets \cite{campos2016evaluation,pang2019devnet}. How to reduce false positives and enhance detection recall rates is one of the most important and yet difficult challenges, particularly for the significant expense of failing to spotting anomalies.
    
    \item \textbf{CH2: Anomaly detection in high-dimensional and/or not-independent data}. Anomalies often exhibit evident abnormal characteristics in a low-dimensional space yet become hidden and unnoticeable in a high-dimensional space. High-dimensional anomaly detection has been a long-standing problem \cite{zimek2012survey}. Performing anomaly detection in a reduced lower-dimensional space spanned by a small subset of original features or newly constructed features is a straightforward solution, \eg, in subspace-based \cite{lazarevic2005fb,liu2012iforest,keller2012hics,pevny2016loda} and feature selection-based methods \cite{pang2017feature,azmandian2012feature,pang2018sparse}. However, identifying intricate (\eg, high-order, nonlinear and heterogeneous) feature interactions and couplings \cite{cao2015coupling} may be essential in high-dimensional data, but it remains a major challenge for anomaly detection. Further, how to guarantee the new feature space preserved proper information for specific detection methods is critical to downstream accurate anomaly detection, but it is challenging due to the aforementioned unknowns and heterogeneities of anomalies. Also, it is challenging to detect anomalies from instances that may be dependent on each other such as by temporal, spatial, graph-based and other interdependency relationships \cite{cao2015coupling,aggarwal2017outlieranalysis,akoglu2015graphsurvey,gupta2013temporaloutliersurvey}.
    
    \item \textbf{CH3: Data-efficient learning of normality/abnormality}. Due to the difficulty and cost of collecting large-scale labeled anomaly data, \textit{fully supervised anomaly detection}  is often impractical as it assumes the availability of labeled training data with both normal and anomaly classes. In the last decade, major research efforts have been focused on \textit{unsupervised anomaly detection} that does not require any labeled training data. However, unsupervised methods do not have any prior knowledge of true anomalies. They rely heavily on their assumption on the distribution of anomalies
    % but fail to work in datasets where their assumption is violated
    . On the other hand, it is often not difficult to collect labeled normal data and some labeled anomaly data. In practice, it is often suggested to leverage such readily accessible labeled data as much as possible \cite{aggarwal2017outlieranalysis}. Thus, utilizing those labeled data to learn expressive representations of normality/abnormality is crucial for accurate anomaly detection. \textit{Semi-supervised anomaly detection}, which assumes a set of labeled 
   normal training data\footnote{There have been some studies that refer the methods trained with purely normal training data to be unsupervised approach. However, this setting is different from the general sense of an unsupervised setting. To avoid unnecessary confusion, following \cite{chandola2009anomaly,aggarwal2017outlieranalysis}, these methods are referred to as semi-supervised methods hereafter.}, is a research direction dedicated to this problem. Another research line is \textit{weakly-supervised anomaly detection} that assumes we have some labels for anomaly classes yet the class labels are partial/incomplete (\ie, they do not span the entire set of anomaly class), inexact (\ie, coarse-grained labels), or inaccurate (\ie, some given labels can be incorrect). Two major challenges are how to learn expressive normality/abnormality representations with a small amount of labeled anomaly data and how to learn detection models that are generalized to novel anomalies uncovered by the given labeled anomaly data. 

    \item \textbf{CH4: Noise-resilient anomaly detection}. Many weakly/semi-supervised anomaly detection methods assume the labeled training data is clean, which can be vulnerable to noisy instances that are mistakenly labeled as an opposite class label. In such cases, we may use unsupervised methods instead, but this fails to utilize the genuine labeled data. Additionally, there often exists large-scale anomaly-contaminated unlabeled data. Noise-resilient models can leverage those unlabeled data for more accurate detection. Thus, the noise here can be either mislabeled data or unlabeled anomalies. The main challenge is that the amount of noises can differ significantly from datasets and noisy instances may be irregularly distributed in the data space.
    
    \item \textbf{CH5: Detection of complex anomalies}. Most of existing methods are for point anomalies, which cannot be used for conditional anomaly and group anomaly since they exhibit completely different behaviors from point anomalies. One main challenge here is to incorporate the concept of conditional/group anomalies into anomaly measures/models. Also, current methods mainly focus on detect anomalies from single data sources, while many applications require the detection of anomalies with multiple heterogeneous data sources, \eg, multidimensional data, graph, image, text and audio data. One main challenge is that some anomalies can be detected only when considering two or more data sources.
    
    % \item \textbf{CH7: Anomaly thresholding}. There are two types of outputs produced from anomaly detection models, including continuous anomaly scores and discrete class labels. \textit{Anomaly scores} are yielded by anomaly scoring methods that aim to assign an anomaly score to each data instance, indicating the degree of being anomalous. A ranking of data instances based on the scores can be derived, with the top-ranked instances reported as anomalies. A cutoff decision threshold is required to obtain the anomalies. \textit{Class labels} are yielded by anomaly classification methods that aim to assign a discrete class label, \eg, \textit{anomalous} or \textit{normal} to each instance. The output of anomaly scores is dominant in current methods. This is because different applications may have different signaling criteria and sensitivity w.r.t. anomalies while the anomaly score output allows an application-dependent cutoff decision threshold for flagging anomalies. However, it is difficult to understand the underlying distribution of the anomaly scores yielded by most methods. As a result, how to properly set this decision threshold is a challenging task, especially for safety/life-critical applications.

    \item \textbf{CH6: Anomaly explanation}. In many safety-critical domains there may be some major risks if anomaly detection models are directly used as black-box models. For example, the \textit{rare} data instances reported as anomalies may lead to possible algorithmic bias against the minority groups presented in the data, such as under-represented groups in fraud detection and crime detection systems. An effective approach to mitigate this type of risk is to have anomaly explanation algorithms that provide straightforward clues about why a specific data instance is identified as anomaly. Human experts can then look into and correct the bias. Providing such explanation can be as important as detection accuracy in some applications. However, most anomaly detection studies focus on detection accuracy only, ignoring the capability of providing explanation of the identified anomalies. 
    % In recent years, there have been some studies \cite{angiulli2009explanation,duan2015explanation,vinh2016explanation,angiulli2017explanation,siddiqui2019explanation} that explore this issue by searching for a subset of features that makes a reported anomaly most abnormal. The abnormal feature selection methods \cite{pang2016feature,azmandian2012feature,pang2017feature} can also be utilized for anomaly explanation purpose. The anomalous feature searching in these methods is often independent from the anomaly detection methods, which may render the explanation less useful as our primary motivation is to understand why the instances are identified as anomalies by specific methods. 
    To derive anomaly explanation from specific detection methods is still a largely unsolved problem, especially for complex models. Developing inherently interpretable anomaly detection models is also crucial, but it remains a main challenge to well balance the model's interpretability and effectiveness.
\end{itemize}

Deep methods enable end-to-end optimization of the whole anomaly detection pipeline, and they also enable the learning of representations specifically tailored for anomaly detection. These two capabilities are crucial to tackle the above six challenges, but traditional methods do not have. Particularly they help largely improve the utilization of labeled normal data or some labeled anomaly data regardless of the data type, reducing the need of large-scale labeled data as in fully-supervised settings (CH2, CH3, CH4, CH5). This subsequently results in more informed models and thus better recall rate (CH1). For the anomaly explanation challenge, although deep methods are often black-box models, they offer options to unify anomaly detection and explanation into single frameworks, resulting in more genuine explanation of the anomalies spotted by specific models (see Section \ref{subsec:interpretablemodel}). Deep methods also excel at learning intricate structures and relations from diverse types of data, such as high-dimensional data, image data, video data, graph data, etc. This capability is important to address various challenges, such as CH1, CH2, CH3, and CH5. Further, they offer many effective and easy-to-use network architectures and principled frameworks to seamlessly learn unified representations of heterogeneous data sources. This empowers the deep models to tackle some key challenges such as CH3 and CH5. Although there are shallow methods for handling those complex data, they are generally substantially weaker and less adaptive than the deep methods. A summary of this discussion is presented in Table \ref{tab:dad_challenges}.

\begin{table*}[htbp]
\centering
\caption{Deep Learning Methods vs. Traditional Methods in Anomaly Detection.}
  \scalebox{0.65}{
    \begin{tabular}{ccccc}
    \hline\hline
    \textbf{Method} & \textbf{End-to-end Optimization} & \textbf{Tailored Representation Learning} & \textbf{Intricate Relation Learning}  & \textbf{Heterogeneity Handling}\\\hline
    Traditional & $\times$ & $\times$ & Weak & Weak\\
    Deep & $\checkmark$ & $\checkmark$ & Strong & Strong\\\hline
    \textbf{Challenges}  & CH1-6 & CH1-6 &  CH1, CH2, CH3, CH5 & CH3, CH5 \\
    \hline\hline
    \end{tabular}
    }
\label{tab:dad_challenges}
\end{table*}

\section{Addressing the Challenges with Deep Anomaly Detection}

\subsection{Preliminaries}

Deep neural networks leverage complex compositions of linear/non-linear functions that can be represented by a computational graph to learn expressive representations \cite{goodfellow2016deep}. Two basic building blocks of deep learning are activation functions and layers. \textit{Activation functions} determine the output of computational graph nodes (\ie, neurons in neural networks) given some inputs. They can be linear or non-linear functions. Some popular activation functions include linear, sigmoid, tanh, ReLU (Rectified Linear Unit) and its variants. A \textit{layer} in neural networks refers to a set of neurons stacked in some forms. Commonly-used layers include fully connected, convolutional \& pooling, and recurrent layers. These layers can be leveraged to build different popular neural networks. For example, multilayer perceptron (MLP) networks are composed by fully connected layers, convolutional neural networks (CNN) are featured by varying groups of convolutional \& pooling layers, and recurrent neural networks (RNN), \eg, vanilla RNN, gated recurrent units (GRU) and long short term memory (LSTM), are built upon recurrent layers. See \cite{goodfellow2016deep} for detailed introduction of these neural networks.

Given a dataset $\mathcal{X} =\{\xvec_1, \xvec_2, \cdots, \xvec_N\}$ with $\xvec_i \in \mathbb{R}^{D}$, let $\mathcal{Z} \in \mathbb{R}^{K}$ ($K \ll N$) be a representation space, then \textbf{deep anomaly detection} aims at learning a feature representation mapping function $\phi(\cdot): \mathcal{X} \mapsto \mathcal{Z}$ or an anomaly score learning function $\tau(\cdot):\mathcal{X} \mapsto \mathbb{R}$ in a way that anomalies can be easily differentiated from the normal data instances in the space yielded by the $\phi$ or $\tau$ function, where both $\phi$ and $\tau$ are a neural network-enabled mapping function with $H \in \mathbb{N}$ hidden layers and their weight matrices $\Theta=\{\mathbf{M}^{1}, \mathbf{M}^{2}, \cdots, \mathbf{M}^{H}\}$. In the case of learning the feature mapping $\phi(\cdot)$, an additional step is required to calculate the anomaly score of each data instance in the new representation space, while $\tau(\cdot)$ can directly infer the anomaly scores with raw data inputs. Larger $\tau$ outputs indicate greater degree of being anomalous.

\subsection{Categorization of Deep Anomaly Detection}

\begin{figure*}[h!]
  \centering
    \includegraphics[width=0.90\textwidth]{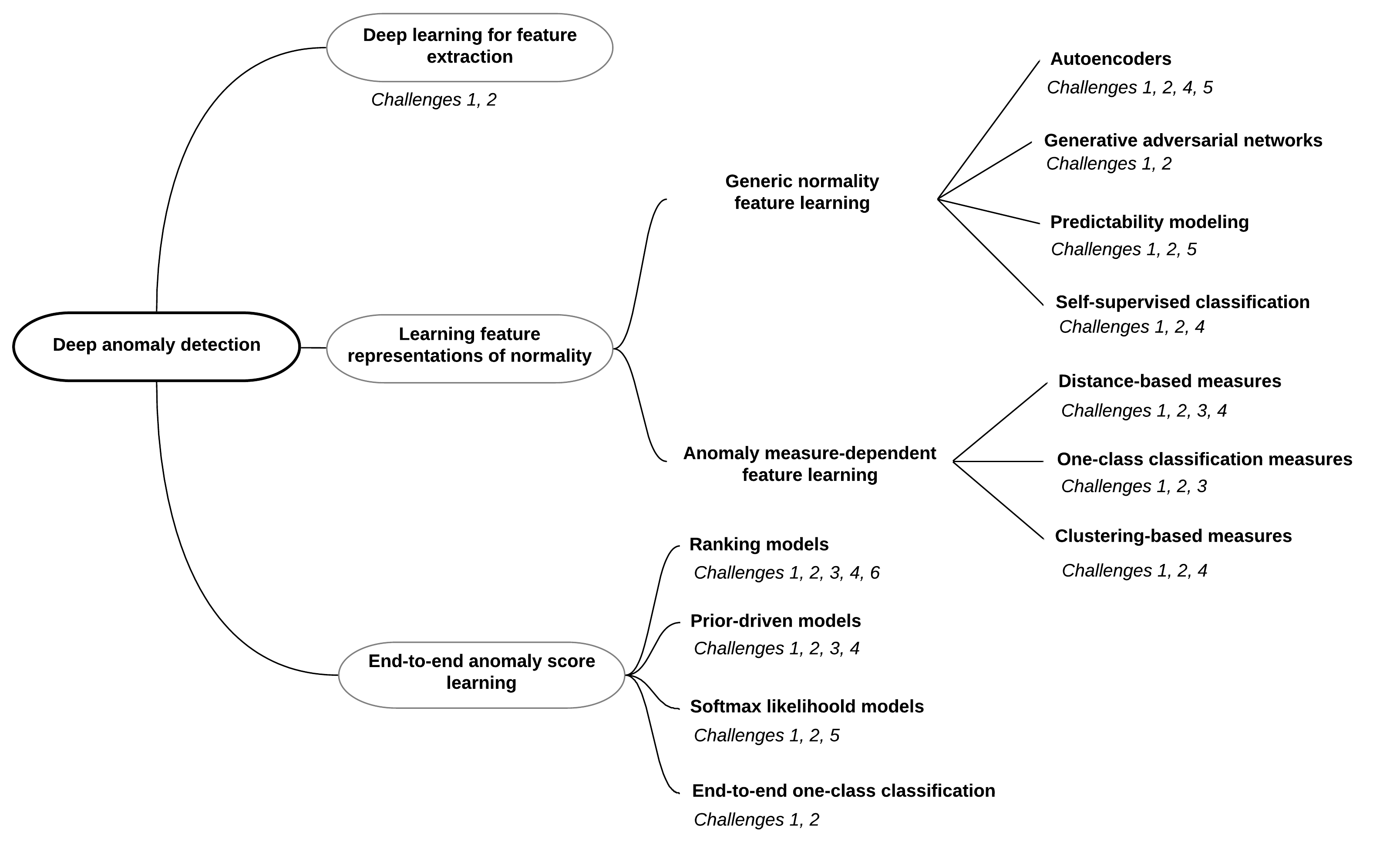}
  \caption{The Proposed Taxonomy of Current Deep Anomaly Detection Techniques. The detection challenges that each category of methods can address are also presented.}
  \label{fig:taxonomy}
\end{figure*}

To have a thorough understanding of the area, we introduce a hierarchical taxonomy to classify deep anomaly detection methods into three main categories and 11 fine-grained categories from the modeling perspective. An overview of the taxonomy of the methods
% , together with the challenges they address,
is shown in Fig. \ref{fig:taxonomy}. Specifically, deep anomaly detection consists of three conceptual paradigms - \textit{Deep Learning for Feature Extraction}, \textit{Learning Feature Representations of Normality}, and \textit{End-to-end Anomaly Score Learning}.

\begin{figure*}[h!]
  \centering
    \includegraphics[width=0.90\textwidth]{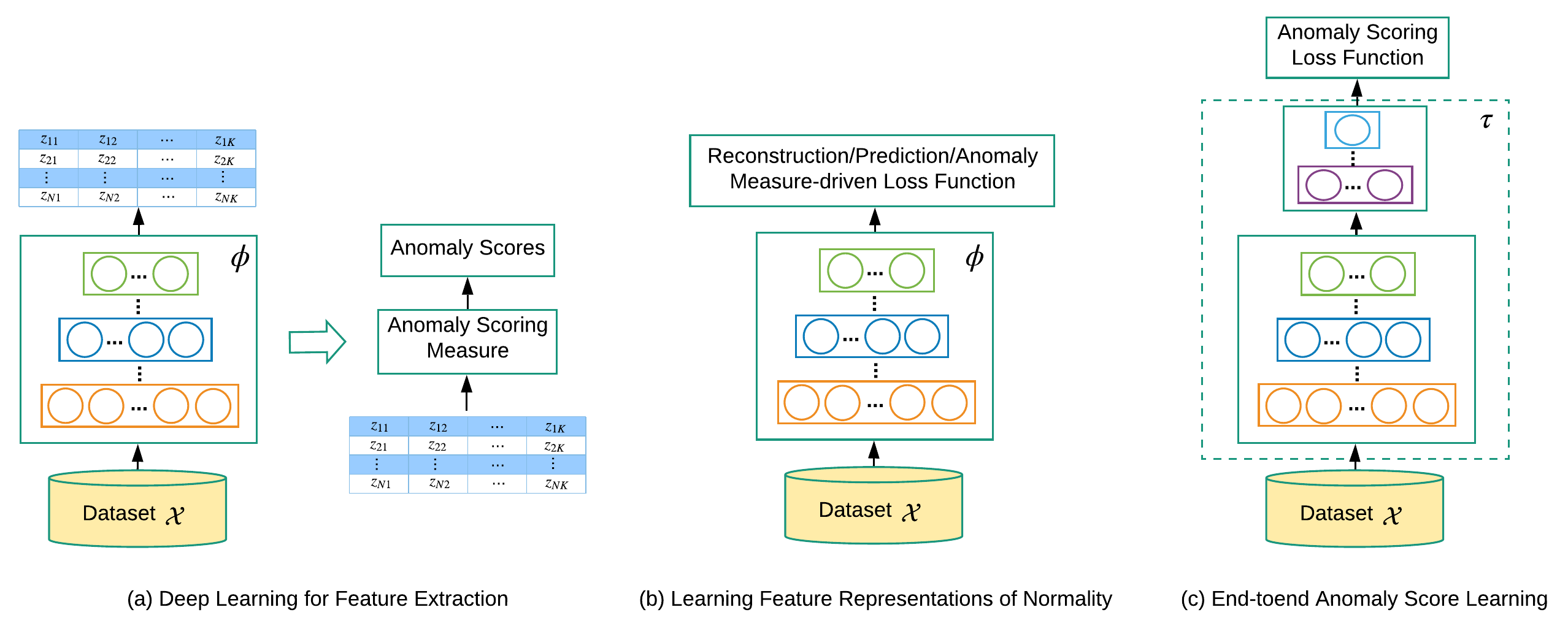}
  \caption{Conceptual Frameworks of Three Main Deep Anomaly Detection Approaches}
  \label{fig:threecategories}
\end{figure*}

The procedure of these three paradigms is presented in Fig. \ref{fig:threecategories}. 
As shown in Fig. \ref{fig:threecategories}(a), deep learning and anomaly detection are fully separated in the first main category (Section \ref{sec:featureextraction}), so deep learning techniques are used as some independent feature extractors only. The two modules are dependent on each other in some form in the second main category (Section \ref{sec:normality}) presented in Fig. \ref{fig:threecategories}(b), with an objective of learning expressive representations of normality. This category of methods can be further divided into two subcategories based on how the representations are learned, \ie, whether using existing shallow anomaly measures (\eg, distance- and clustering-based measures) to guide the learning or not. These two subcategories encompass seven fine-grained categories of methods, with each category taking a different approach to formulate its objective function. The two modules are fully unified in the third main category (Section \ref{sec:anomalyscore}) presented in Fig. \ref{fig:threecategories}(c), in which the methods are dedicated to learning anomaly scores via neural networks in an end-to-end fashion. This category is further broken down into four subcategories based on the formulation of the anomaly scoring learning. In the following three sections we review these three paradigms in detail.
% the methods in each of these three categories in detail and discuss how they address some of the aforementioned challenges.

\section{Deep Learning for Feature Extraction}\label{sec:featureextraction}
This category of methods aims at leveraging deep learning to extract low-dimensional feature representations from high-dimensional and/or non-linearly separable data for downstream anomaly detection. The feature extraction and the anomaly scoring are fully disjointed and independent from each other. Thus, the deep learning components work purely as dimensionality reduction only. Formally, the approach can be represented as

\begin{equation}\label{eqn:featureextraction}
    \mathbf{z} = \phi(\mathbf{x}; \Theta),
\end{equation}
where $\phi:\mathcal{X} \mapsto \mathcal{Z}$ is a deep neural network-based feature mapping function, with $\mathcal{X}\in \mathbb{R}^{D}$, $\mathcal{Z} \in \mathbb{R}^{K}$ and normally $D \gg K$. An anomaly scoring method $f$ that has no connection to the feature mapping $\phi$ is then applied onto the new space to calculate anomaly scores.

Compared to the dimension reduction methods that are popular in anomaly detection, such as principal component analysis (PCA) \cite{scholkopf1997kernelpca,zou2006sparsepca,candes2011robustpca} and random projection \cite{li2006pr,pevny2016loda,pang2018repen}, deep learning techniques have been demonstrating substantially better capability in extracting semantic-rich features and non-linear feature relations \cite{bengio2013representation,goodfellow2016deep}.

\textit{Assumptions}. The feature representations extracted by deep learning models preserve the discriminative information that helps separate anomalies from normal instances.

One research line is to directly uses popular pre-trained deep learning models, such as AlexNet \cite{krizhevsky2012alexnet}, VGG \cite{simonyan2015vgg} and ResNet \cite{he2016resnet}, to extract low-dimensional features. This line is explored in anomaly detection in complex high-dimensional data such as image data and video data. One interesting work of this line is the unmasking framework for online anomaly detection \cite{tudor2017unmasking}. The key idea is to iteratively train a binary classifier to separate one set of video frames from its subsequent video frames in a sliding window, with the most discriminant features removed in each iteration step. This is analogous to an unmasking process. The framework assumes the first set of video frames as normal and evaluates its separability from its subsequent video frames. Thus, the training classification accuracy is expected to be high if the subsequent video frames are abnormal, and low otherwise. 
% The unmasking is an anomaly scoring process, with the change of the training accuracy used to define the anomaly scores. 
Clearly the power of the unmasking framework relies heavily on the quality of the features, so it is essential to have quality features to represent the video frames. The VGG model pre-trained on the ILSVRC benchmark \cite{russakovsky2015imagenet} is shown to be effective to extract expressive appearance features for this purpose \cite{tudor2017unmasking}. In \cite{liu2018unmasking}, the masking framework is formulated as a two-sample test task to understand its theoretical foundation. They also show that using features extracted from a dynamically updated sampling pool of video frames is found to improve the performance of the framework. Additionally, similar to many other tasks, the feature representations extracted from the deep models pre-trained on a source dataset can be transferred to fine-tune anomaly detectors on a target dataset. As shown in \cite{andrews2016transfer}, one-class support vector machines (SVM) can be first initialized with the VGG models pre-trained on ILSVRC and then fine-tuned to improve anomaly classification on the MNIST data \cite{lecun1998mnist}. Similarly, the ResNet models pre-trained on MNIST can empower abnormal event detection in various video surveillance datasets \cite{zhou2019anomalynet,pang2020ranking}.

Another research line in this category is to explicitly train a deep feature extraction model rather than a pre-trained model for the downstream anomaly scoring \cite{xu2015videoae,erfani2016featureextractor,ionescu2019featureextraction,yu2018featureextraction}. Particularly, in \cite{xu2015videoae}, three separate autoencoder networks are trained to learn low-dimensional features for respective appearance, motion, and appearance-motion joint representations for video anomaly detection. An ensemble of three one-class SVMs is independently trained on each of these learned feature representations to perform anomaly scoring. Similar to \cite{xu2015videoae}, a linear one-class SVM is used to enable anomaly detection on low-dimensional representations of high-dimensional tabular data yielded by deep belief networks (DBNs) \cite{erfani2016featureextractor}. Instead of one-class SVM, unsupervised classification approaches are used in \cite{ionescu2019featureextraction} to enable anomaly scoring in the projected space. Specially, they first cluster the low-dimensional features of video frames yielded by convolutional autoencoders, and then treat the cluster labels as pseudo class labels to perform one-vs-the-rest classification. The classification probabilities are used to define frame-wise anomaly scores. Similar approaches can also be found in graph anomaly detection \cite{yu2018featureextraction}, in which unsupervised clustering-based anomaly measures are used in the latent representation space to calculate the abnormality of graph vertices or edges. To learn expressive representations of graph vertices, the vertex representations are optimized by minimizing autoencoder-based reconstruction loss and pairwise distances of neighbored graph vertices, taking one-hot encoding of graph vertices as input.

\textit{Advantages}. The advantages of this approach are as follows. (i) A large number of state-of-the-art (pre-trained) deep models and off-the-shelf anomaly detectors are readily available. (ii) Deep feature extraction offers more powerful dimensionality reduction than popular linear methods. (iii) It is easy-to-implement given the public availability of the deep models and detection methods.

\textit{Disadvantages}. Their disadvantages are as follows. (i) The fully disjointed feature extraction and anomaly scoring often lead to suboptimal anomaly scores. (ii) Pre-trained deep models are typically limited to specific types of data.

\textit{Challenges Targeted}. This category of methods projects high-dimensional/non-independent data onto substantially lower-dimensional space, enabling existing anomaly detection methods to work on simpler data space. The lower-dimensional space often helps reveal hidden anomalies and reduces false positives (\textbf{CH2}). However, it should be noted that these methods may not preserve sufficient information for anomaly detection as the data projection is fully decoupled with anomaly detection. In addition, this approach allows us to leverage multiple types of features and learn semantic-rich detection models (\eg, various predefined image/video features in \cite{xu2015videoae,tudor2017unmasking,ionescu2019featureextraction}), which also helps reduce false positives (\textbf{CH1}).

\section{Learning Feature Representations of Normality}\label{sec:normality}

% This section reviews the models from the perspective of normality learning. 
The methods in this category couple feature learning with anomaly scoring in some ways, rather than fully decoupling these two modules as in the last section. These methods generally fall into two groups: generic feature learning and anomaly measure-dependent feature learning. 
% Below we discuss these two types of approaches in detail.

\subsection{Generic Normality Feature Learning}\label{subsec:genericnormality}
This category of methods learns the representations of data instances by optimizing a generic feature learning objective function that is not primarily designed for anomaly detection, but the learned representations can still empower the anomaly detection since they are forced to capture some key underlying data regularities. Formally, this framework can be represented as

\begin{gather}
    \{\Theta^*, \mathbf{W}^*\} = \argmin_{\Theta,\ \mathbf{W}} \sum_{\xvec \in \mathcal{X}}\ell\Big(\psi\big(\phi(\mathbf{x};\Theta);\mathbf{W}\big)\Big) \label{eqn:genericfeature1},\\
    s_{\mathbf{x}} = f(\mathbf{x}, \phi_{\Theta^*}, \psi_{\mathbf{W}^*}) \label{eqn:genericfeature2},
\end{gather}
where $\phi$ maps the original data onto the representation space $\mathcal{Z}$, $\psi$ parameterized by $\mathbf{W}$ is a surrogate learning task that operates on the $\mathcal{Z}$ space and is dedicated to enforcing the learning of underlying data regularities, $\ell$ is a loss function relative to the underlying modeling approach, and $f$ is a scoring function that utilizes $\phi$ and $\psi$ to calculate the anomaly score $s$.

This approach include methods driven by several perspectives, including data reconstruction, generative modeling, predictability modeling and self-supervised classification. Both predictability modeling and self-supervised classification are built upon self-supervised learning approaches, but they have different assumptions, advantages and flaws, and thus they are reviewed separately.

% Therefore, the reconstruction/prediction errors of data instances can be directly used to define anomaly scores. Based on the approach of the prediction, below we review three types of methods in this direction: data compression, generative modeling and predictability modeling methods.

\subsubsection{Autoencoders}\label{subsubsec:ae}

This type of approach aims to learn some low-dimensional feature representation space on which the given data instances can be well reconstructed. This is a widely-used technique for data compression or dimension reduction \cite{hinton2006ae,jiang2014dimensionalityreductionae,theis2017dimensionalityreductionae}. The heuristic for using this technique in anomaly detection is that the learned feature representations are enforced to learn important regularities of the data to minimize reconstruction errors; anomalies are difficult to be reconstructed from the resulting representations and thus have large reconstruction errors.
% . As a result, the reconstruction error for each instance can be naturally defined as anomaly score, since normal instances normally can be better represented by the compressed feature space than anomalies. 

% Larger reconstruction errors indicate higher likelihood of being anomalies. 

\textit{Assumptions}. Normal instances can be better restructured from compressed space than anomalies.
% Two basic assumptions of this approach are as follows:
% \begin{itemize}
%     \item Normal data instances can be better restructured from low-dimensional feature representations than anomalies.
%     \item Training data contains only (or an overwhelming amount of) normal data.
% \end{itemize}

Autoencoder (AE) networks are the commonly-used techniques in this category. An AE is composed of an encoding network and an decoding network. The encoder maps the original data onto low-dimensional feature space, while the decoder attempts to recover the data from the projected low-dimensional space. The parameters of these two networks are learned with a reconstruction loss function. A bottleneck network architecture is often used to obtain low-dimensional representations than the original data, which forces the model to retain the information that is important in reconstructing the data instances. To minimize the overall reconstruction error, the retained information is required to be as much relevant as possible to the dominant instances, \eg, the normal instances. As a result, the data instances such as anomalies that deviate from the majority of the data are poorly reconstructed. The data reconstruction error can therefore be directly used as anomaly score. The basic formulation of this approach is given as follows.
\begin{gather}\label{eqn:ae}
    \mathbf{z} = \phi_e(\mathbf{x}; \Theta_e),\; \hat{\mathbf{x}} = \phi_d(\mathbf{z}; \Theta_d),\\
   \{\Theta_e^*, \Theta_d^*\} = \argmin_{\Theta_e,\Theta_d} \sum_{\xvec \in \mathcal{X}}\big\|\mathbf{x}-\phi_d\big(\phi_e(\mathbf{x};\Theta_e); \Theta_d\big)\big\|^2,\\
    s_{\mathbf{x}} = \big\|\xvec - \phi_d\big(\phi_e(\mathbf{x};\Theta_e^*); \Theta_d^*\big)\big\|^2,
\end{gather}
where $\phi_e$ is the encoding network with the parameters $\Theta_e$ and $\phi_d$ is the decoding network with the parameters $\Theta_d$. The encoder and the decoder can share the same weight parameters to reduce parameters and regularize the learning. $s_{\xvec}$ is a reconstruction error-based anomaly score of $\xvec$.

Several types of regularized autoencoders have been introduced to learn richer and more expressive feature representations \cite{makhzani2014sparseae,vincent2010denoisingae,rifai2011contractiveae,doersch2016variationalae}. Particularly, sparse AE is trained in a way that encourages sparsity in the activation units of the hidden layer, \eg, by keeping the top-$K$ most active units \cite{makhzani2014sparseae}. Denoising AE \cite{vincent2010denoisingae} aims at learning representations that are robust to small variations by learning to reconstruct data from some predefined corrupted data instances rather than original data. Contractive AE \cite{rifai2011contractiveae} takes a step further to learn feature representations that are robust to small variations of the instances around their neighbors. This is achieved by adding a penalty term based on the Frobenius norm of the Jacobian matrix of the encoder's activations. Variational AE \cite{doersch2016variationalae} instead introduces regularization into the representation space by encoding data instances using a prior distribution over the latent space, preventing overfitting and ensuring some good properties of the learned space for enabling generation of meaningful data instances.

AEs are easy-to-implement and have straightforward intuitions in detecting anomalies. As a result, they have been widely explored in the literature. Replicator neural network \cite{hawkins2002autoencoder} is the first piece of work in exploring the idea of data reconstruction to detect anomalies, with experiments focused on static multidimensional/tabular data. The Replicator network is built upon a feed-forward multi-layer perceptron with three hidden layers. It uses \textit{parameterized} hyperbolic tangent activation functions to obtain different activation levels for different input values, which helps discretize the intermediate representations into some predefined bins. As a result, the hidden layers naturally cluster the data instances into a number of groups, enabling the detection of clustered anomalies. After this work there have been a number of studies dedicated to further enhance the performance of this approach. For instance, RandNet \cite{chen2017ensembleae} further enhances the basic AEs by learning an ensemble of AEs. In RandNet, a set of independent AEs are trained, with each AE having some randomly selected constant dropout connections. An adaptive sampling strategy is used by exponentially increasing the sample size of the mini-batches. RandNet is focused on tabular data. The idea of autoencoder ensembles is extended to time series data in \cite{kieu2019aeensemble}. Motivated by robust principal component analysis (RPCA), RDA \cite{zhou2017autoencoder} attempts to improve the robustness of AEs by iteratively decomposing the original data into two subsets, normal instance set and anomaly set. This is achieved by adding a sparsity penalty $\ell_1$ or grouped penalty $\ell_{2,1}$ into its RPCA-alike objective function to regularize the coefficients of the anomaly set.

AEs are also widely leveraged to detect anomalies in data other than tabular data, such as sequence data \cite{lu2017sequenceae}, graph data \cite{ding2019graphae} and image/video data \cite{xu2015videoae}. In general, there are two types of adaptions of AEs to those complex data. The most straightforward way is to follow the same procedure as the conventional use of AEs by adapting the network architecture to the type of input data, such as CNN-AE \cite{hasan2016convae,zhang2019convolutionalae}, LSTM-AE \cite{malhotra2016lstmae}, Conv-LSTM-AE \cite{luo2017convolutionlstmae} and GCN (graph convolutional network)-AE \cite{ding2019graphae}. This type of AEs embeds the encoder-decoder scheme into the full procedure of these methods. Another type of AE-based approaches is to first use AEs to learn low-dimensional representations of the complex data and then learn to predict these learned representations. The learning of AEs and representation prediction is often two separate steps. These approaches are different from the first type of approaches in that the prediction of representations are wrapped around the low-dimensional representations yielded by AEs. For example, in \cite{lu2017sequenceae}, denoising AE is combined with RNNs to learn normal patterns of multivariate sequence data, in which a denoising AE with two hidden layers is first used to learn representations of multidimensional data inputs in each time step and a RNN with a single hidden layer is then trained to predict the representations yielded by the denoising AE. A similar approach is also used for detecting acoustic anomalies \cite{marchi2015lstmae}, in which a more complex RNN, bidirectional LSTMs, is used. 

\textit{Advantages}. The advantages of data reconstruction-based methods are as follows. (i) The idea of AEs is straightforward and generic to different types of data. (ii) Different types of powerful AE variants can be leveraged to perform anomaly detection.

\textit{Disadvantages}. Their disadvantages are as follows. (i) The learned feature representations can be biased by infrequent regularities and the presence of outliers or anomalies in the training data. (ii) The objective function of the data reconstruction is designed for dimension reduction or data compression, rather than anomaly detection. As a result, the resulting representations are a generic summarization of underlying regularities, which are not optimized for detecting irregularities.

\textit{Challenges Targeted}. Different types of neural network layers and architectures can be used under the AE framework, allowing us to detect anomalies in high-dimensional data, as well as non-independent data such as attributed graph data \cite{ding2019graphae} and multivariate sequence data \cite{marchi2015lstmae,lu2017sequenceae} (\textbf{CH2}). These methods may reduce false positives over traditional methods built upon handcrafted features if the learned representations are more expressive (\textbf{CH1}). AEs are generally vulnerable to data noise presented in the training data as they can be trained to remember those noise, leading to severe overfitting and small reconstruction errors of anomalies. The idea of RPCA may be used into AEs to train more robust detection models \cite{zhou2017autoencoder} (\textbf{CH4}).

% , including static multidimensional/tabular data \cite{hawkins2002autoencoder,chen2017ensembleae}, graph data \cite{ding2019graphae}, sequence data \cite{lu2017sequenceae} and image/video data \cite{xu2015videoae}

\subsubsection{Generative Adversarial Networks}\label{subsubsec:generative}

GAN-based anomaly detection emerges quickly as one popular deep anomaly detection approach after its early use in \cite{schlegl2017generative}. This approach generally aims to learn a latent feature space of a generative network $G$ so that the latent space well captures the normality underlying the given data. Some form of residual between the real instance and the generated instance are then defined as anomaly score.

\textit{Assumption}. Normal data instances can be better generated than anomalies from the latent feature space of the generative network in GANs.

One of the early methods is AnoGAN \cite{schlegl2017generative}. The key intuition is that, given any data instances $\mathbf{x}$, it aims to search for an instance $\mathbf{z}$ in the learned latent feature space of the generative network $G$ so that the corresponding generated instance $G(\mathbf{z})$ and $\mathbf{x}$ are as similar as possible. Since the latent space is enforced to capture the underlying distribution of training data, anomalies are expected to be less likely to have highly similar generated counterparts than normal instances. Specifically, a GAN is first trained with the following conventional objective:
\begin{equation}\label{eqn:anogan}
    \min_{G} \max_{D} V(D, G) = \mathbb{E}_{\mathbf{x} \sim p_{\mathcal{X}} } \big[\log D(\mathbf{x})\big] + 
    \mathbb{E}_{\mathbf{z} \sim p_{\mathcal{Z}} } \bigg[\log\Big(1- D\big(G(\mathbf{z})\big)\Big)\bigg],
\end{equation}
where $G$ and $D$ are respectively the generator and discriminator networks parameterized by $\Theta_G$ and $\Theta_D$ (the parameters are omitted for brevity), and $V$ is the value function of the two-player minimax game. After that, for each $\mathbf{x}$, to find its best $\mathbf{z}$, two loss functions - residual loss and discrimination loss - are used to guide the search. The residual loss is defined as
\begin{equation}\label{eqn:residualloss}
    \ell_{R}(\mathbf{x}, \mathbf{z}_{\gamma})= \big\|\mathbf{x}-G(\mathbf{z}_{\gamma})\big\|_1,
\end{equation}
while the discrimination loss is defined based on the feature matching technique \cite{salimans2016improvedgan}:
\begin{equation}\label{eqn:featurematching}
    \ell_{\mathit{fm}}(\mathbf{x}, \mathbf{z}_{\gamma})=\big\|h(\mathbf{x})-h\big(G(\mathbf{z}_{\gamma})\big)\big\|_1,
\end{equation}
where $\gamma$ is the index of the search iteration step and $h$ is a feature mapping from an intermediate layer of the discriminator. The search starts with a randomly sampled $\mathbf{z}$, followed by updating $\mathbf{z}$ based on the gradients derived from the overall loss $(1-\alpha) \ell_{R}(\mathbf{x}, \mathbf{z}_{\gamma}) + \alpha \ell_{\mathit{fm}}(\mathbf{x}, \mathbf{z}_{\gamma})$, where $\alpha$ is a hyperparameter. Throughout this search process, the parameters of the trained GAN are fixed; the loss is only used to update the coefficients of $\mathbf{z}$ for the next iteration. The anomaly score is accordingly defined upon the similarity between $\mathbf{x}$ and $\mathbf{z}$ obtained at the last step $\gamma^*$:

\begin{equation}\label{eqn:anoganscore}
    s_{\mathbf{x}}= (1-\alpha) \ell_{R}(\mathbf{x}, \mathbf{z}_{\gamma^*}) + \alpha \ell_{\mathit{fm}}(\mathbf{x}, \mathbf{z}_{\gamma^*}).
\end{equation}

One main issue with AnoGAN is the computational inefficiency in the iterative search of $\mathbf{z}$. One way to address this issue is to add an extra network that learns the mapping from data instances onto the latent space, \ie, an inverse of the generator, resulting in methods like EBGAN \cite{zenati2018generative} and fast AnoGAN \cite{schlegl2019generative}. These two methods share the same spirit. Here we focus on EBGAN that is built upon the bi-directional GAN (BiGAN) \cite{donahue2017bigan}. Particularly, BiGAN has an encoder $E$ to map $\mathbf{x}$ to $\mathbf{z}$ in the latent space, and simultaneously learn the parameters of $G$, $D$ and $E$. Instead of discriminating $\mathbf{x}$ and $G(\mathbf{z})$, BiGAN aims to discriminate the pair of instances $(\mathbf{x}, E(\mathbf{x}))$ from the pair $(G(\mathbf{z}),\mathbf{z})$:

% V(G,E,D) = 
\begin{multline}\label{eqn:ebgan}
    \min_{G,E} \max_{D} \; \mathbb{E}_{\mathbf{x} \sim p_{\mathcal{X}} } \Big[\mathbb{E}_{\mathbf{z}\sim p_{E}(\cdot|\mathbf{x})}\log \big[D(\mathbf{x},\mathbf{z})\big]\Big] + 
    \mathbb{E}_{\mathbf{z} \sim p_{\mathcal{Z}} } \bigg[ \mathbb{E}_{\mathbf{x} \sim p_{G}(\cdot|\mathbf{z})}\Big[\log \big(1- D(\mathbf{x},\mathbf{z})\big)\Big]\bigg],
\end{multline}

After the training, inspired by Eq. (\ref{eqn:anoganscore}) in AnoGAN, EBGAN defines the anomaly score as:
\begin{equation}
   s_{\mathbf{x}}= (1-\alpha) \ell_G(\mathbf{x}) + \alpha \ell_{D}(\mathbf{x}),
\end{equation}
where $\ell_G(\mathbf{x}) = \big\|\mathbf{x}-G\big(E(\mathbf{x})\big)\big\|_1$ and $\ell_{D}(\mathbf{x}) = \big\|h\big(\mathbf{x},E(\mathbf{x})\big)-h\big(G\big(E(\mathbf{x})\big),E(\mathbf{x})\big)\big\|_1$. This eliminates the need to iteratively search $\mathbf{z}$ in AnoGAN. EBGAN is extended to a method called ALAD \cite{zenati2018generativeicdm} by adding two more discriminators, with one discriminator trying to discriminate the pair $(\mathbf{x}, \mathbf{x})$ from $(\mathbf{x}, G(E(\mathbf{x})))$ and another one trying to discriminate the pair $(\mathbf{z}, \mathbf{z})$ from $(\mathbf{z}, E(G(\mathbf{z})))$.

GANomaly \cite{akcay2018generative} further improves the generator over the previous work by changing the generator network to an encoder-decoder-encoder network and adding two more extra loss functions. The generator can be conceptually represented as: $\mathbf{x} \xrightarrow{G_{E}} \mathbf{z} \xrightarrow{G_{D}} \hat{\mathbf{x}} \xrightarrow{E} \hat{\mathbf{z}}$, in which $G$ is a composition of the encoder $G_{E}$ and the decoder $G_{D}$. In addition to the commonly used feature matching loss:
\begin{equation}
    \ell_{\mathit{fm}}= \mathbb{E}_{\mathbf{x} \sim p_{\mathcal{X}}} \big\|h(\mathbf{x}) - h\big(G(\mathbf{x})\big)\big\|_{2},
\end{equation}
the generator includes a contextual loss and an encoding loss to generate more realistic instances:

\begin{equation}\label{eqn:contextloss}
    \ell_{\mathit{con}}= \mathbb{E}_{\mathbf{x} \sim p_{\mathcal{X}}} \big\|\mathbf{x} - G(\mathbf{x})\big\|_{1},    
\end{equation}

\begin{equation}\label{eqn:encloss}
    \ell_{\mathit{enc}}= \mathbb{E}_{\mathbf{x} \sim p_{\mathcal{X}}} \big\|G_{E}(\mathbf{x}) - E\big(G(\mathbf{x})\big)\big\|_{2}.   
\end{equation}
The contextual loss in Eq. (\ref{eqn:contextloss}) enforces the generator to consider the contextual information of the input $\mathbf{x}$ when generating $\hat{\mathbf{x}}$. The encoding loss in Eq. (\ref{eqn:encloss}) helps the generator to learn how to encode the features of the generated instances. The overall loss is then defined as 
\begin{equation}
    \ell = \alpha \ell_{\mathit{fm}} + \beta \ell_{\mathit{con}} + \gamma \ell_{\mathit{enc}},
\end{equation}
where $\alpha$, $\beta$ and $\gamma$ are the hyperparameters to determine the weight of each individual loss. Since the training data contains mainly normal instances, the encoders $G$ and $E$ are optimized towards the encoding of normal instances, and thus, the anomaly score can be defined as
\begin{equation}
    s_{\mathbf{x}} = \big\|G_{E}(\mathbf{x}) - E\big(G(\mathbf{x})\big)\big\|_{1},
\end{equation}
in which $s_{\mathbf{x}}$ is expected to be large if $\mathbf{x}$ is an anomaly.

There have been a number of other GANs introduced over the years such as Wasserstein GAN \cite{arjovsky2017wassersteingan} and Cycle GAN \cite{zhu2017cyclegan}. They may be used to further enhance the anomaly detection performance of the above methods, such as replacing the standard GAN with Wasserstein GAN \cite{schlegl2019generative}. Another relevant research line is to adversarially learn end-to-end one-class classification models, which is categorized into the end-to-end anomaly score learning framework and discussed in Section \ref{subsec:oneclasse2e}.

\textit{Advantages}. The advantages of these methods are as follows. (i) GANs have demonstrated superior capability in generating realistic instances, especially on image data, empowering the detection of abnormal instances that are poorly reconstructed from the latent space. (ii) A large number of existing GAN-based models and theories \cite{creswell2018generative} may be adapted for anomaly detection.

\textit{Disadvantages}. Their disadvantages are as follows. (i) The training of GANs can suffer from multiple problems, such as failure to converge and mode collapse \cite{metz2017modecollapse}, which leads to to large difficulty in training GANs-based anomaly detection models. (ii) The generator network can be misled and generates data instances out of the manifold of normal instances, especially when the true distribution of the given dataset is complex or the training data contains unexpected outliers. (iii) The GANs-based anomaly scores can be suboptimal since they are built upon the generator network with the objective designed for data synthesis rather than anomaly detection.

\textit{Challenges Targeted}. Similar to AEs, GAN-based anomaly detection is able to detect high-dimensional anomalies by examining the reconstruction from the learned low-dimensional latent space (\textbf{CH2}). When the latent space preserves important anomaly discrimination information, it helps improve detection accuracy over that in the original data space (\textbf{CH1}). 
% These GAN-based methods are also dependent on the reconstruction error-based anomaly scores, so they share similar characteristics as the AE-based approach in addressing the CH5 challenge.

\subsubsection{Predictability Modeling} \label{subsec:selfprediction}

Predictability modeling-based methods learn feature representations by predicting the current data instances using the representations of the previous instances within a temporal window as the context. In this section data instances are referred to as individual elements in a sequence, \eg, video frames in a video sequence. This technique is widely used for sequence representation learning and prediction \cite{sutskever2014predictability,mathieu2016predictability,hsieh2018predictability,liao2018predictability}. To achieve accurate predictions, the representations are enforced to capture the temporal/sequential and recurrent dependence within a given sequence length. Normal instances are normally adherent to such dependencies well and can be well predicted, whereas anomalies often violate those dependencies and are unpredictable. Therefore, the prediction errors
% , \eg, measured by mean squared errors or likelihood values, 
can be used to define the anomaly scores. 

\textit{Assumption}. Normal instances are temporally more predictable than anomalies.

This research line is popular in video anomaly detection \cite{liu2018predictabilitycvpr,ye2019predictabilitymm,abati2019predictabilitycvpr}. Video sequence involves complex high-dimensional spatial-temporal features. Different constraints over appearance and motion features are needed in the prediction objective function to ensure a faithful prediction of video frames. This deep anomaly detection approach is initially explored in \cite{liu2018predictabilitycvpr}. Formally, given a video sequence with consecutive $t$ frames $\mathbf{x}_1,\mathbf{x}_2,\cdots,\mathbf{x}_t$, then the learning task is to use all these frames to generate a future frame $\hat{\mathbf{x}}_{t+1}$ so that $\hat{\mathbf{x}}_{t+1}$ is as close as possible to the ground truth $\mathbf{x}_{t+1}$. Its general objective function can be formulated as

\begin{equation}\label{eqn:predictability}
    \alpha \ell_{\mathit{pred}}\big(\hat{\mathbf{x}}_{t+1}, \mathbf{x}_{t+1}\big) + \beta \ell_{\mathit{adv}}\big(\hat{\mathbf{x}}_{t+1}\big),
\end{equation}
where $\hat{\mathbf{x}}_{t+1}=\psi\big(\phi(\mathbf{x}_1,\mathbf{x}_2,\cdots,\mathbf{x}_t;\Theta);\mathbf{W}\big)$, $\ell_{\mathit{pred}}$ is the frame prediction loss measured by mean squared errors, $\ell_{\mathit{adv}}$ is an adversarial loss. The popular network architecture named U-Net \cite{ronneberger2015unet} is used to instantiate the $\psi$ function for the frame generation. $\ell_{\mathit{pred}}$ is composed by a set of three separate losses that respectively enforce the closeness between $\hat{\mathbf{x}}_{t+1}$ and $\mathbf{x}_{t+1}$ in three key image feature descriptors: intensity, gradient and optical flow. $\ell_{\mathit{adv}}$ is due to the the use of adversarial training to enhance the image generation. After training, for a given video frame  $\mathbf{x}$, a normalized Peak Signal-to-Noise Ratio \cite{mathieu2016predictability} based on the prediction difference $||\mathbf{x}_{i}-\hat{\mathbf{x}}_{i}||_2$ is used to define the anomaly score. Under the same framework, an additional autoencoder-based reconstruction network is added in \cite{ye2019predictabilitymm} to further refine the predicted frame quality, which helps to enlarge the anomaly score difference between normal and abnormal frames. 

Another research line in this direction is based on the autoregressive models \cite{gregor2014predictabilityar} that assume each element in a sequence is linearly dependent on the previous elements. The autoregressive models are leveraged in \cite{abati2019predictabilitycvpr} to estimate the density of training samples in a latent space, which helps avoid the assumption of a specific family of distributions. Specifically, given $\mathbf{x}$ and its latent space representation $\mathbf{z}=\phi(\mathbf{x};\Theta)$, the autoregressive model factorizes $p(\mathbf{z})$ as
\begin{equation}
    p(\mathbf{z}) = \prod_{j=1}^{K}p(z_j|z_{1:j-1}),
\end{equation}
where $z_{1:j-1}=\{z_1,z_2,\cdots,z_{j-1}\}$, $p(z_j|z_{1:j-1})$ represents the probability mass function of $z_j$ conditioned on all the previous instances $z_{1:j-1}$ and $K$ is the dimensionality size of the latent space. The objective in \cite{abati2019predictabilitycvpr} is to jointly learn an autoencoder and a density estimation network $\psi(\mathbf{z};\mathbf{W})$ equipped with autoregressive network layers. The overall loss can be represented as
\begin{equation}
    L = \mathbb{E}_{\mathbf{x}}\bigg[ \big\|\mathbf{x} -\phi_d\big(\phi_e(\mathbf{x};\Theta_e);\Theta_d\big)\big\|_2 - \lambda \log \Big( \psi(\mathbf{z};\mathbf{W})\Big) \bigg],
\end{equation}
where the first term is a reconstruction error measured by MSE while the second term is an autoregressive loss measured by the log-likelihood of the representation under an estimated conditional probability density prior. Minimizing this loss enables the learning of the features that are common and easily predictable. At the evaluation stage, the reconstruction error and the log-likelihood are combined to define the anomaly score.

\textit{Advantages}. The advantages of this category of methods are as follows. (i) A number of sequence learning techniques can be adapted and incorporated into this approach. (ii) This approach enables the learning of different types of temporal and spatial dependencies.

\textit{Disadvantages}. Their disadvantages are as follows. (i) This approach is limited to anomaly detection in sequence data. (ii) The sequential predictions can be computationally expensive. (iii) The learned representations may suboptimal for anomaly detection as its underlying objective is for sequential predictions rather than anomaly detection.

\textit{Challenges Targeted}. This approach is particularly designed to learn expressive temporally-dependent low-dimensional representations, which helps address the false positives of anomaly detection in high-dimensional and/or temporal datasets (\textbf{CH1 \& CH2}). The prediction here is conditioned on some elapsed temporal instances, so this category of methods is able to detect temporal context-based conditional anomalies (\textbf{CH5}). 

\subsubsection{Self-supervised Classification}\label{subsubsec:consistency}
This approach learns representations of normality by building self-supervised classification models and identifies instances that are inconsistent to the classification models as anomalies. This approach is rooted in traditional methods based on cross-feature analysis or feature models \cite{huang2003consistency,noto2012consisitency,tenenboim2013consistency}. These shallow methods evaluate the normality of data instances by their consistency to a set of predictive models, with each model learning to predict one feature based on the rest of the other features. The consistency of a test instance can be measured by the average prediction results \cite{huang2003consistency}, the log loss-based surprisal \cite{noto2012consisitency}, or the majority voting of binary decisions \cite{tenenboim2013consistency} given the classification/regression models across all features. Unlike these studies that focus on tabular data and build the feature models using the original data, deep consistency-based anomaly detection focuses on image data and builds the predictive models by using feature transformation-based augmented data. To effectively discriminate the transformed instances, the classification models are enforced to learn features that are highly important to describe the underlying patterns of the instances presented in the training data. Therefore, normal instances generally have stronger agreements with the classification models.

\textit{Assumptions}. Normal instances are more consistent to self-supervised classifiers than anomalies.

This approach is initially explored in \cite{golan2018featuretrasnformation}. To build the predictive models, different compositions of geometric transformation operations, including horizontal flipping, translations, and rotations, is first applied to normal training images. A deep multi-class classification model is trained on the augmented data, treating data instances with a specific transformation operation comes from the same class, \ie, synthetic class. At inference, test instances are augmented with each of transformation compositions, and their normality score is defined by an aggregation of all softmax classification scores to the augmented test instance. Its loss function is defined as 

\begin{equation}\label{eqn:cons}
    L_{\mathit{cons}}= \mathit{CE}\Big(\psi\big(\mathbf{z}_{T_{j}}; \mathbf{W}\big), \mathbf{y}_{T_{j}}\Big),
\end{equation}
where $\mathbf{z}_{T_{j}}=\phi\big(T_{j}({\mathbf{x}}); \Theta\big)$ is a low-dimensional feature representation of instance $\mathbf{x}$ augmented by the transformation operation type $T_{j}$, $\psi$ is a multi-class classifier parameterized with $\mathbf{W}$, $\mathbf{y}_{T_{j}}$ is a one-hot encoding of the synthetic class for instances augmented using the transformation operation $T_{j}$, and $\mathit{CE}$ is a standard cross-entropy loss function.

By minimizing Eq. (\ref{eqn:cons}), we obtain the representations that are optimized for the classifier $\psi$. We then can apply the feature learner $\phi(\cdot, \Theta^*)$ and the classifier $\psi(\cdot, \mathbf{W}^*)$ to obtain a classification score for each test instance augmented with a transformation operation $T_{j}$. The classification scores of each test instance w.r.t. different $T_{j}$ are then aggregated to compute the anomaly score. To achieve that, the classification scores conditioned on each $T_{j}$ is assumed to follow a Dirichlet distribution in \cite{golan2018featuretrasnformation} to estimate the consistency of the test instance to the classification model $\psi$; a simple average of the classification scores associated with different $T_{j}$ also works well. 
% . Actually, as shown in \cite{golan2018featuretrasnformation}, a simple average of the classification scores associated with different $T_{j}$ works similarly well as the Dirichlet-based anomaly score.

A semi-supervised setting, \ie, training data contains normal instances only, is assumed in \cite{golan2018featuretrasnformation}. A similar idea is explored in the unsupervised setting in \cite{wang2019featuretransformation}, with the transformation sets containing four transformation operations, \ie, rotation, flipping, shifting and path re-arranging. Two key insights revealed in \cite{wang2019featuretransformation} is that (i) the gradient magnitude induced by normal instances is normally substantially larger than outliers during the training of such self-supervised multi-class classification models; and (ii) the network updating direction is also biased towards normal instances. As a result of these two properties, normal instances often have stronger agreement with the classification model than anomalies. Three strategies of using the classification scores to define the anomaly scores are evaluated, including average prediction probability, maximum prediction probability, and negative entropy across all prediction probabilities \cite{wang2019featuretransformation}. Their results show that the negative entropy-based anomaly scores perform generally better than the other two strategies.

\textit{Advantages}. The advantages of deep consistency-based methods are as follows. (i) They work well in both the unsupervised and semi-supervised settings. (ii) Anomaly scoring is grounded by some intrinsic properties of gradient magnitude and its updating.

\textit{Disadvantages}. Their disadvantages are as follows. (i) The feature transformation operations are often data-dependent. The above transformation operations are applicable to image data only. 
% Different transformation operations need to be explored to adapt this approach to other types of data. 
(ii) Although the classification model is trained in an end-to-end manner, the consistency-based anomaly scores are derived upon the classification scores rather than an integrated module in the optimization, and thus they may be suboptimal.

\textit{Challenges Targeted}. The expressive low-dimensional representations of normality this approach learns help detect anomalies better than in the original high-dimensional space (\textbf{CH1 \& CH2}). Due to some intrinsic differences between anomalies and normal instances presented in the self-supervised classifiers, this approach is also able to work in an unsupervised setting \cite{wang2019featuretransformation}, demonstrating good robustness to anomaly contamination in the training data (\textbf{CH4}).

\subsection{Anomaly Measure-dependent Feature Learning}\label{subsec:measuredependent}

Anomaly measure-dependent feature learning aims at learning feature representations that are specifically optimized for one particular existing anomaly measure. Formally, the framework for this group of methods can be represented as
\begin{gather}
    \{\Theta^*, \mathbf{W}^*\} = \argmin_{\Theta,\ \mathbf{W}} \sum_{\mathbf{x} \in \mathcal{X}}\ell\Big(f\big(\phi(\mathbf{x};\Theta);\mathbf{W}\big)\Big),\\
    s_{\xvec} = f\big(\phi(\xvec;\Theta^*);\mathbf{W}^*\big),
\end{gather}
where $f$ is an existing anomaly scoring measure operating on the representation space. Note that whether $f$ may involve trainable parameters $\mathbf{W}$ or not is dependent on the anomaly measure used. Different from the generic feature learning approach as in Eqs. (\ref{eqn:genericfeature1}-\ref{eqn:genericfeature2}) that calculates anomaly scores based on some heuristics after obtaining the learned representations, this research line incorporates an existing anomaly measure $f$ into the feature learning objective function to optimize the feature representations specifically for $f$. Below we review representation learning specifically designed for three types of popular anomaly measures, including distance-based measure, one-class classification measure and clustering-based measure.
% so that $f\big(\phi(\mathbf{x}_{i};\Theta)\big) > f\big(\phi(\mathbf{x}_{j};\Theta)\big)$, where $\mathbf{x}_{i}$ is an anomaly, $\mathbf{x}_{j}$ is a normal instance, $\phi$ is a feature learner, and $f$ is an existing anomaly scoring measure operated on the representation space.
\subsubsection{Distance-based Measure}\label{subsubsec:distancemeasure}

Deep distance-based anomaly detection aims to learn feature representations that are specifically optimized for a specific type of distance-based anomaly measures. Distance-based methods are straightforward and easy-to-implement. There have been a number of effective distance-based anomaly measures introduced, \eg, DB outliers \cite{knorr1999finding,knorr2000distance}, $k$-nearest neighbor distance \cite{ramaswamy2000efficientdistance,ramaswamy2000knndistance}, average $k$-nearest neighbor distance \cite{angiulli2002averagedistance}, relative distance \cite{zhang2009distance}, and random nearest neighbor distance \cite{sugiyama2013rapiddistance,pang2015lesinn}. One major limitation of these traditional distance-based anomaly measures is that they fail to work effectively in high-dimensional data due to the curse of dimensionality. Since deep distance-based anomaly detection techniques project data onto low-dimensional space before applying the distance measures, it can well overcome this limitation.

\textit{Assumption}. Anomalies are distributed far from their closest neighbors while normal instances are located in dense neighborhoods.

This approach is first explored in \cite{pang2018repen}, in which the random neighbor distance-based anomaly measure \cite{sugiyama2013rapiddistance,pang2015lesinn} is leveraged to drive the learning of low-dimensional representations out of ultrahigh-dimensional data. The key idea is that the representations are optimized so that the nearest neighbor distance of pseudo-labeled anomalies in random subsamples is substantially larger than that of pseudo-labeled normal instances. The pseudo labels are generated by some off-the-shelf anomaly detectors. Let $\mathcal{S} \in \mathcal{X}$ be a subset of data instances randomly sampled from the dataset $\mathcal{X}$, $\mathcal{A}$ and $\mathcal{N}$ respectively be the pseudo-labeled anomaly and normal instance sets, with $\mathcal{X} = \mathcal{A} \cup \mathcal{N}$ and $\emptyset = \mathcal{A} \cap \mathcal{N}$, its loss function is built upon the hinge loss function \cite{rosasco2004loss}:

\begin{equation}\label{eqn:distanceobj}
    L_{\mathit{query}} = \frac{1}{|\mathcal{X}|}\sum_{\mathbf{x} \in \mathcal{A}, \mathbf{x}^{\prime} \in \mathcal{N}} \max\big\{0, m + f(\mathbf{x}^{\prime},\mathcal{S};\Theta) - f(\mathbf{x},\mathcal{S};\Theta) \big\},
\end{equation}
where $m$ is a predefined constant for the margin between two distances yielded by $f(\mathbf{x},\mathcal{S};\Theta)$, which is a random nearest neighbor distance function operated in the representation space:
\begin{equation}\label{eqn:randomdistance}
    f(\mathbf{x},\mathcal{S};\Theta) = \min_{\mathbf{x}^\prime \in \mathcal{S}} \; \big\|\phi(\mathbf{x};\Theta), \phi(\mathbf{x}^\prime;\Theta)\big\|_2.
\end{equation}

% $L_{\mathit{query}}$ is a hinge loss function augmented by the random nearest neighbor distance-based anomaly measure defined in Eq. (\ref{eqn:randomdistance}). 
Minimizing the loss in Eq. (\ref{eqn:distanceobj}) guarantees that the random nearest neighbor distance of anomalies are at least $m$ greater than that of normal instances in the $\phi$-projected representation space. At the evaluation stage, the random distance in Eq. (\ref{eqn:randomdistance}) is used directly to obtain the anomaly score for each test instance. Following this approach, we might also derive similar representation learning tailored for other distance-based measures by replacing Eq. (\ref{eqn:randomdistance}) with the other measures, such as the $k$-nearest neighbor distance \cite{ramaswamy2000knndistance} or the average $k$-nearest neighbor distance \cite{angiulli2002averagedistance}. However, these measures are significantly more computationally costly than Eq. (\ref{eqn:randomdistance}). Thus, one major challenging for such adaptions would be the prohibitively high computational cost.

Compared to \cite{pang2018repen} that requires to query the nearest neighbor distances in random data subsets, inspired by \cite{burda2019curiosityb}, a simpler idea explored in \cite{wang2020distance} uses the distance between optimized representations and randomly projected representations of the same instances to guide the representation learning. The objective of the method is as follows
\begin{equation}\label{eqn:distillation}
    \Theta^* = \argmin_{\Theta} \sum_{\mathbf{x} \in \mathcal{X}} f\big(\phi(\mathbf{x};\Theta), \phi^{\prime}(\mathbf{x})\big),
\end{equation}
where $\phi^{\prime}$ is a random mapping function that is instantiated by the neural network used in $\phi$ with fixed random weights, $f$ is a measure of distance between the two representations of the same data instance. As discussed in \cite{burda2019curiosityb}, solving Eq. (\ref{eqn:distillation}) is equivalent to have a knowledge distillation from a random neural network and helps learn the frequency of different underlying patterns in the data. However, Eq. (\ref{eqn:distillation}) ignores the relative proximity between data instances and is sensitive to the anomalies presented in the data. As shown in \cite{wang2020distance}, such proximity information may be learned by a pretext task, in which we aim to predict the distance between random instance pairs. A boosting process can also be used to iteratively filter potential anomalies and build robust detection models. At the evaluation stage, $f(\phi(\mathbf{x};\Theta^*), \phi^{\prime}(\mathbf{x}))$ is used to compute the anomaly scores.

\textit{Advantages}. The advantages of this category of methods are as follows. (i) The distance-based anomalies are straightforward and well defined with rich theoretical supports in the literature. Thus, deep distance-based anomaly detection methods can be well grounded due to the strong foundation built in previous relevant work. (ii) They work in low-dimensional representation spaces and can effectively deal with high-dimensional data that traditional distance-based anomaly measures fail. (iii) They are able to learn representations specifically tailored for themselves.

\textit{Disadvantages}. Their disadvantages are as follows. (i) The extensive computation involved in most of distance-based anomaly measures may be an obstacle to incorporate distance-based anomaly measures into the representation learning process. (ii) Their capabilities may be limited by the inherent weaknesses of the distance-based anomaly measures.

\textit{Challenges Targeted}. This approach is able to learn low-dimensional representations tailored for existing distance-based anomaly measures, addressing the notorious curse of dimensionality in distance-based detection \cite{zimek2012survey} (\textbf{CH1 \& CH2}). As shown in \cite{pang2018repen}, an adapted triplet loss can be devised to utilize a few labeled anomaly examples to learn more effective normality representations (\textbf{CH3}). Benefiting from pseudo anomaly labeling, the methods \cite{pang2018repen,wang2020distance} are also robust to potential anomaly contamination and work effectively in the fully unsupervised setting (\textbf{CH4}).

\subsubsection{One-class Classification-based Measure}\label{subsubsec:oneclass}
This category of methods aims to learn feature representations customized to subsequent one-class classification-based anomaly detection. One-class classification is referred to as the problem of learning a description of a set of data instances to detect whether new instances conform to the training data or not. It is one of the most popular approaches for anomaly detection \cite{moya1993oneclass,scholkopf2001oneclasssvm,tax2004svdd,roth2005oneclasskf}. Most one-class classification models are inspired by Support Vector Machines (SVM) \cite{cortes1995svm}, such as the two widely-used one-class models: one-class SVM (or $v$-SVC) \cite{scholkopf2001oneclasssvm} and Support Vector Data Description (SVDD) \cite{tax2004svdd}. One main research line here is to learn representations that are specifically optimized for these traditional one-class classification models. This is the focus of this section. Another line is to learn an end-to-end adversarial one-class classification model, which will be discussed in Section \ref{subsec:oneclasse2e}.

\textit{Assumption}. All normal instances come from a single (abstract) class and can be summarized by a compact model, to which anomalies do not conform.

There are a number of studies dedicated to combine one-class SVM with neural networks \cite{wu2019oneclasssvm,nguyen2018oneclasssvm,chalapathy2018oneclasssvm}. Conventional one-class SVM is to learn a hyperplane that maximize a margin between training data instances and the origin. The key idea of deep one-class SVM is to learn the one-class hyperplane from the neural network-enabled low-dimensional representation space rather than the original input space. Let $\mathbf{z}=\phi(\mathbf{x}; \Theta)$, then a generic formulation of the key ideas in \cite{wu2019oneclasssvm,nguyen2018oneclasssvm,chalapathy2018oneclasssvm} can be represented as
\begin{equation}\label{eqn:oneclasssvm}
    \min_{r, \Theta, \mathbf{w}} \frac{1}{2}||\Theta||^2 + \frac{1}{vN}\sum_{i=1}^{N} \max\big\{0, r - \mathbf{w}^{\intercal}\mathbf{z}_i\big\} - r,
\end{equation}
where $r$ is the margin parameter, $\Theta$ are the parameters of a representation network, and $\mathbf{w}^{\intercal}\mathbf{z}$ (\ie, $\mathbf{w}^{\intercal}\phi(\mathbf{x}; \Theta)$) replaces the original dot product $\big<\mathbf{w}, \Phi(\mathbf{x})\big>$ that satisfies $k(\mathbf{x}_i, \mathbf{x}_j)=\big<\Phi(\mathbf{x}_i),\Phi(\mathbf{x}_j)\big>$. Here $\Phi$ is a RKHS (Reproducing Kernel Hilbert Space) associated mapping and $k(\cdot,\cdot)$ is a kernel function; $v$ is a hyperparameter that can be seen as an upper bound of the fraction of the anomalies in the training data. Any instances that have $r - \mathbf{w}^{\intercal}\mathbf{z}_i > 0$ can be reported as anomalies.

This formulation brings two main benefits: (i) it can leverage (pretrained) deep networks to learn more expressive features for downstream anomaly detection, and (iii) it also helps remove the computational expensive pairwise distance computation in the kernel function. As shown in \cite{wu2019oneclasssvm,nguyen2018oneclasssvm}, the reconstruction loss in AEs can be added into Eq. (\ref{eqn:oneclasssvm}) to enhance the expressiveness of representations $\mathbf{z}$. As shown in \cite{rahimi2008fourier}, many kernel functions can be approximated with random Fourier features. Thus, before $\mathbf{w}^{\intercal}\mathbf{z}$, some form of random mapping $h$ may be applied to $\mathbf{z}$ to generate Fourier features, resulting in $\mathbf{w}^{\intercal}h(\mathbf{z})$, which may further improve one-class SVM models. Another research line studies deep models for SVDD \cite{ruff2018deepsvdd,ruff2020deepsvdd}. SVDD aims to learn a minimum hyperplane characterized by a center $\mathbf{c}$ and a radius $r$ so that the sphere contains all training data instances, \ie,
% \begin{equation}
%     ||\mathbf{x}_i - \mathbf{c}||^2 \leq r^2, \; \forall i.
% \end{equation}

% SVDD minimizes the volume of the sphere by minimizing $r^2$ with a set of slack variables $\xi$:
\begin{gather}
    \min_{r, \mathbf{c}, \xi} \; r^2 + \frac{1}{vN}\sum_{i=1}^N \xi_i\\
    \text{s.t.} \; ||\Phi(\mathbf{x}_i)-\mathbf{c}||^2 \leq r^2 + \xi_i, \ \xi_i \geq 0, \; \forall i.
\end{gather}

Similar to deep one-class SVM, deep SVDD \cite{ruff2018deepsvdd} also aims to leverage neural networks to map data instances into the sphere of minimum volume, and then employs the hinge loss function to guarantee the margin between the sphere center and the projected instances. The feature learning and the SVDD objective can then be jointly trained by minimizing the following loss:
\begin{equation}
    \min_{r, \Theta} \; r^2+\frac{1}{vN}\sum_{i=1}^{N} \max\{0, ||\phi(\mathbf{x}_i;\Theta)-\mathbf{c}||^2-r^2\} +\frac{\lambda}{2} ||\Theta||^2.
\end{equation}
This assume the training data contains a small proportion of anomaly contamination in the unsupervised setting. In the semi-supervised setting, the loss function can be simplified as
\begin{equation}\label{eqn:oneclassdeepsvdd}
    \min_{\Theta}\; \frac{1}{N}||\phi(\mathbf{x}_i;\Theta)-\mathbf{c}||^2+\frac{\lambda}{2} ||\Theta||^2,
\end{equation}
which directly minimizes the mean distance between the representations of training data instances and the center $\mathbf{c}$. Note that including $\mathbf{c}$ as trainable parameters in Eq. (\ref{eqn:oneclassdeepsvdd}) can lead to trivial solutions. It is shown in \cite{ruff2018deepsvdd} that $\mathbf{c}$ can be fixed as the mean of the feature representations yielded by performing a single initial forward pass. Deep SVDD can also be further extended to address another semi-supervised setting where a small number of both labeled normal instances and anomalies are available \cite{ruff2020deepsvdd}. The key idea is to minimize the distance of labeled normal instances to the center while at the same time maximizing the distance of known anomalies to the center. This can be achieved by adding $\sum_{j=1}^{M}\big( ||\phi(\mathbf{x}_j^{\prime};\Theta)-\mathbf{c}||^2\big)^{y_j}$ into Eq. (\ref{eqn:oneclassdeepsvdd}), where $\mathbf{x}_j^{\prime}$ is a labeled instance, $y_j=+1$ when it is a normal instance and $y_j=-1$ otherwise.

\textit{Advantages}. The advantages of this category of methods are as follows. (i) The one-class classification-based anomalies are well studied in the literature and provides a strong foundation of deep one-class classification-based methods. (ii) The representation learning and one-class classification models can be unified to learn tailored and more optimal representations. (iii) They free the users from manually choosing suitable kernel functions in traditional one-class models.

\textit{Disadvantages}. Their disadvantages are as follows. (i) The one-class models may work ineffectively in datasets with complex distributions within the normal class.  (ii) The detection performance is dependent on the one-class classification-based anomaly measures.

\textit{Challenges Targeted}. This category of methods enhances detection accuracy by learning lower-dimensional representation space optimized for one-class classification models (\textbf{CH1 \& CH2}). A small number of labeled normal and abnormal data can be leveraged by these methods \cite{ruff2020deepsvdd} to learn more effective one-class description models, which can not only detect known anomalies but also novel classes of anomaly (\textbf{CH3}).  

\subsubsection{Clustering-based Measure}\label{subsubsec:clustering}
Deep clustering-based anomaly detection aims at learning representations so that anomalies are clearly deviated from the clusters in the newly learned representation space. The task of clustering and anomaly detection is naturally tied with each other, so there have been a large number of studies dedicated to using clustering results to define anomalies, \eg, cluster size \cite{jiang2001clustering}, distance to cluster centers \cite{he2003clustering}, distance between cluster centers \cite{jiang2006clustering}, and cluster membership \cite{schubert2017clustering}. Gaussian mixture model-based anomaly detection \cite{mahadevan2010clustering,emmott2013clustering} is also included into this category due to some of its intrinsic relations to clustering, \eg, the likelihood fit in the Gaussian mixture model (GMM) corresponds to an aggregation of the distances of data instances to the centers of the Gaussian clusters/components \cite{aggarwal2017outlieranalysis}.

\textit{Assumptions}. Normal instances have stronger adherence to clusters than anomalies.

Deep clustering, which aims to learn feature representations tailored for a specific clustering algorithm, is the most critical component of this anomaly detection method. A number of studies have explored this problem in recent years \cite{tian2014clustering,xie2016clustering,yang2016clustering,dilokthanakul2017clustering,ghasedi2017clustering,caron2018clustering,yang2019spectralclustering}. The main motivation is due to the fact that the performance of clustering methods is highly dependent on the input data. Learning feature representations specifically tailored for a clustering algorithm can well guarantee its performance on different datasets \cite{aljalbout2018clusteringsurvey}. In general, there are two key intuitions here: (i) good representations enables better clustering and good clustering results can provide effective supervisory signals to representation learning; and (ii) representations that are optimized for one clustering algorithm is not necessarily useful for other clustering algorithms due to the difference of the underlying assumptions made by the clustering algorithms. 
% The targeted clustering algorithm can be $k$-means clustering \cite{xie2016clustering,guo2017clustering,caron2018clustering,jabi2019clustering}, spectral clustering \cite{tian2014clustering,yang2019spectralclustering}, agglomerative clustering \cite{yang2016clustering}. 

The deep clustering methods typically consist of two modules: performing clustering in the forward pass and learning representations using the cluster assignment as pseudo class labels in the backward pass. Its loss function is often the most critical part, which can be generally formulated as
\begin{equation}\label{eqn:clustering}
    \alpha \ell_{\mathit{clu}}\Big(f\big(\phi(\mathbf{x};\Theta); \mathbf{W}\big), y_{x}\Big) + \beta \ell_{\mathit{aux}}(\mathcal{X}),
\end{equation}
where $\ell_{\mathit{clu}}$ is a clustering loss function, within which $\phi$ is the feature learner parameterized by $\Theta$, $f$ is a clustering assignment function parameterized by $\mathbf{W}$ and $y_{x}$ represents pseudo class labels yielded by the clustering; $\ell_{\mathit{aux}}$ is a non-clustering loss function used to enforce additional constrains on the learned representations; and $\alpha$ and $\beta$ are two hyperparameters to control the importance of the two losses. $\ell_{\mathit{clu}}$ can be instantiated with a $k$-means loss \cite{xie2016clustering,caron2018clustering}, a spectral clustering loss \cite{tian2014clustering,yang2019spectralclustering}, an agglomerative clustering loss \cite{yang2016clustering}, or a GMM loss \cite{dilokthanakul2017clustering}, enabling the representation learning for the targeted clustering algorithm. $\ell_{\mathit{aux}}$ is often instantiated with an autoencoder-based reconstruction loss \cite{ghasedi2017clustering,yang2019spectralclustering} to learn robust and/or local structure preserved representations.
% , or to prevent collapsing clusters.

After the deep clustering, the cluster assignments in the resulting $f$ function can then be utilized to compute anomaly scores based on \cite{jiang2001clustering,he2003clustering,jiang2006clustering,schubert2017clustering}. However, it should be noted that the deep clustering may be biased by anomalies if the training data is anomaly-contaminated. Therefore, the above methods can be applied to the semi-supervised setting where the training data is composed by normal instances only. In the unsupervised setting, some extra constrains are required in $\ell_{\mathit{clu}}$ and/or $\ell_{\mathit{aux}}$ to eliminate the impact of potential anomalies.

The aforementioned deep clustering methods are focused on learning optimal clustering results. Although their clustering results are applicable to anomaly detection, the learned representations may not be able to well capture the abnormality of anomalies. It is important to utilize clustering techniques to learn representations so that anomalies have clearly weaker adherence to clusters than normal instances. Some promising results for this type of approach are shown in \cite{zong2018autoencoder,liao2018clustering}, in which they aim to learn representations for a GMM-based model with the representations optimized to highlight anomalies. The general formation of these two studies is similar to Eq. (\ref{eqn:clustering}) with $\ell_{\mathit{clu}}$ and $\ell_{\mathit{aux}}$ respectively specified as a GMM loss and an autoencoder-based reconstruction loss, but to learn deviated representations of anomalies, they concatenate some handcrafted features based on the reconstruction errors with the learned features of the autoencoder to optimize the combined features together. Since the reconstruction error-based handcrafted features capture the data normality, the resulting representations are more optimal for anomaly detection than that yielded by other deep clustering methods.

\textit{Advantages}. The advantages of deep clustering-based methods are as follows. (i) A number of deep clustering methods and theories can be utilized to support the effectiveness and theoretical foundation of anomaly detection. (ii) Compared to traditional clustering-based methods, deep clustering-based methods learn specifically optimized representations that help spot the anomalies easier than on the original data, especially when dealing with intricate data sets.

\textit{Disadvantages}. Their disadvantages are as follows. (i) The performance of anomaly detection is heavily dependent on the clustering results. (ii) The clustering process may be biased by contaminated anomalies in the training data, which in turn leads to less effective representations. 
% (iii) The feature representations are suboptimal to anomaly detection if the objective function of the deep clustering is not properly designed to highlight anomalies.

\textit{Challenges Targeted}. The clustering-based anomaly measures are applied to newly learned low-dimensional representations of data inputs; when the new representation space preserves sufficient discrimination information, the deep methods can achieve better detection accuracy than that in the original data space (\textbf{CH1 \& CH2}). Some clustering algorithms are sensitive to outliers, so the deep clustering and the subsequent anomaly detection can be largely misled when the given data is contaminated by anomalies. Deep clustering using handcrafted features from the reconstruction errors of autoencoders \cite{zong2018autoencoder} may help learn more robust models w.r.t. the contamination (\textbf{CH4}).

\section{End-to-end Anomaly Score Learning}\label{sec:anomalyscore}

This research line aims at learning scalar anomaly scores in an end-to-end fashion. Compared to anomaly measure-dependent feature learning, the anomaly scoring in this type of approach is not dependent on existing anomaly measures; it has a neural network that directly learns the anomaly scores. Novel loss functions are often required to drive the anomaly scoring network. Formally, this approach aims at learning an end-to-end anomaly score learning network: $\tau(\cdot; \Theta):\mathcal{X} \mapsto \mathbb{R}$. The underlying framework can be represented as
\begin{gather}\label{eqn:scorelearning1}
    \Theta^* = \argmin_{\Theta} \sum_{\xvec \in \mathcal{X}} \ell\big( \tau(\mathbf{x};\Theta) \big),\\
    s_{\xvec} = \tau(\mathbf{x};\Theta^*)\label{eqn:scorelearning2}.
\end{gather}
% which is trainable in an end-to-end fashion.

% that can be seen as a sequential composite of a feature representation learning network $\phi(\cdot; \Theta_{t}): \mathcal{X} \mapsto \mathcal{Z}$ and a neural anomaly score learning network $\eta(\cdot; \Theta_{s}): \mathcal{Z} \mapsto \mathbb{R}$, in which $\Theta=\{\Theta_{t}, \Theta_{s}\}$

Unlike those methods in Section \ref{subsec:genericnormality} that use some sort of heuristics to calculate anomaly scores after obtaining the learned representations, the methods in this category simultaneously learn the feature representations and anomaly scores. This greatly optimizes the anomaly scores and/or anomaly ranking. In this perspective they share some similarities as the methods in Section \ref{subsec:measuredependent}. However, the anomaly measure-dependent feature learning methods are often limited by the inherent disadvantages of the incorporated anomaly measures, whereas the methods here do not have such weakness; they also represent two completely different directions of designing the models: one focuses on how to synthesize existing anomaly measures and neural network models, while another focuses on devising novel loss functions for direct anomaly score learning.

Below we review four main approaches in this category: ranking models, prior-driven models, softmax likelihood models and end-to-end one-class classification models. The key to this framework is to incorporate order or discriminative information into the anomaly scoring network. 
% These four approaches represent four different perspectives to design this network.

\subsection{Ranking Models}\label{subsec:ranking}

This group of methods aims to directly learn a ranking model, such that data instances can be sorted based on an observable ordinal variable associated with the absolute/relative ordering relation of the abnormality. The anomaly scoring neural network is driven by the observable ordinal variable.

\textit{Assumptions}. There exists an observable ordinal variable that captures some data abnormality.

One research line of this approach is to devise ordinal regression-based loss functions to drive the anomaly scoring neural network \cite{pang2019ranking,pang2020ranking}. In \cite{pang2020ranking}, a self-trained deep ordinal regression model is introduced to directly optimize the anomaly scores for unsupervised video anomaly detection. Particularly, it assumes an observable ordinal variable $\mathbf{y} =\{c_1,c_2\}$ with $c_1 > c_2$, let $\tau(\mathbf{x};\Theta) = \eta(\phi(\mathbf{x};\Theta_{t});\Theta_s)$, $\mathcal{A}$ and $\mathcal{N}$ respectively be pseudo anomaly and normal instance sets and $\mathcal{G}=\mathcal{A} \cup \mathcal{N}$, then the objective function is formulated as
\begin{equation}\label{eqn:ordinalregression}
    \argmin_{\Theta} \sum_{\mathbf{x} \in 
    \mathcal{G} } \ell\big(\tau(\mathbf{x};\Theta), \ysubx\big),
\end{equation}
where $\ell(\cdot,\cdot)$ is a MSE/MAE-based loss function and $\ysubx = c_1 \;,  \forall \mathbf{x} \in \mathcal{A}$ and $\ysubx = c_2 \;, \forall \mathbf{x} \in \mathcal{N}$. Here $y$ takes two scalar ordinal values only, so it is a two-class ordinal regression. 

The end-to-end anomaly scoring network takes $\mathcal{A}$ and $\mathcal{N}$ as inputs and learns to optimize the anomaly scores such that the data inputs of similar behaviors as those in $\mathcal{A}$ ($\mathcal{N}$) receive large (small) scores as close $c_1$ ($c_2$) as possible, resulting in larger anomaly scores assigned to anomalous frames than normal frames. Due to the superior capability of capturing appearance features of image data, ResNet-50 \cite{he2016resnet} is used to specify the feature network $\phi$, followed by the anomaly scoring network $\eta$ built with a fully connected two-layer neural network. $\eta$ consists of a hidden layer with 100 units and an output layer with a single linear unit. Similar to \cite{pang2018repen}, $\mathcal{A}$ and $\mathcal{N}$ are initialized by some existing anomaly measures. The anomaly scoring model is then iteratively updated and enhanced in a self-training manner. The MAE-based loss function is employed in Eq. (\ref{eqn:ordinalregression}) to reduce the negative effects brought by false pseudo labels in $\mathcal{A}$ and $\mathcal{N}$. 

Different from \cite{pang2020ranking} that addresses an unsupervised setting, a weakly-supervised setting is assumed in \cite{pang2019ranking,sultani2018ranking}. A very small number of labeled anomalies, together with large-scale unlabeled data, is assumed to be available during training in \cite{pang2019ranking}. To leverage the known anomalies, the anomaly detection problem is formulated as a pairwise relation prediction task. Specifically, a two-stream ordinal regression network is devised to learn the relation of randomly sampled pairs of data instances, \ie, to discriminate whether the instance pair contains two labeled anomalies, one labeled anomaly, or just unlabeled data instances. Let $\mathcal{A}$ be the small labeled anomaly set, $\mathcal{U}$ be the large unlabeled dataset and $\mathcal{X}= \mathcal{A} \cup \mathcal{U}$, $\mathcal{P} = \big\{ \{\mathbf{x}_{i}, \mathbf{x}_{j}, y_{\mathbf{x}_i\mathbf{x}_j}\} \,| \, \mathbf{x}_{i}, \mathbf{x}_{j} \in \mathcal{X} \; \text{and}\; y_{\mathbf{x}_i\mathbf{x}_j} \in \mathbb{N} \big\}$ is first generated. Here $\mathcal{P}$ is a set of random instance pairs with \textit{synthetic} ordinal class labels, where $\mathbf{y}=\{y_{\mathbf{x}_{a_i}\mathbf{x}_{a_j}}, y_{\mathbf{x}_{a_i}\mathbf{x}_{u_i}}, y_{\mathbf{x}_{u_i}\mathbf{x}_{u_j}}\}$ is an ordinal variable. The synthetic label $y_{\mathbf{x}_{a_i}\mathbf{x}_{u_i}}$ means an ordinal value for any instance pairs with the instances $\mathbf{x}_{a_i}$ and $\mathbf{x}_{u_i}$ respectively sampled from $\mathcal{A}$ and $\mathcal{U}$. $y_{\mathbf{x}_{a_i}\mathbf{x}_{a_j}} > y_{\mathbf{x}_{a_i}\mathbf{x}_{u_i}} > y_{\mathbf{x}_{u_i}\mathbf{x}_{u_j}}$ is predefined such that the pairwise prediction task is equivalent to anomaly score learning. The method can then be formally framed as

\begin{equation}\label{eqn:pairordinal}
    \Theta^* = \argmin_{\Theta} \; \frac{1}{|\mathcal{P}|}\sum_{\{\mathbf{x}_{i},\mathbf{x}_{j}, y_{ij}\} \in \mathcal{P}} \Big|y_{\mathbf{x}_i\mathbf{x}_j} - \tau\big((\mathbf{x}_{i},\mathbf{x}_{j});\Theta\big)\Big|,
\end{equation}
which is trainable in an end-to-end fashion. By minimizing Eq. (\ref{eqn:pairordinal}), the model is optimized to learn larger anomaly scores for the pairs of two anomalies than the pairs with one anomaly or none. At inference, each test instance is paired with instances from $\mathcal{A}$ or $\mathcal{U}$ to obtain the anomaly scores.

The weakly-supervised setting in \cite{sultani2018ranking} addresses frame-level video anomaly detection, but only video-level class labels are available during training, \ie, a video is normal or contains abnormal frames somewhere - we do not know which specific frames are anomalies. A multiple instance learning (MIL)-based ranking model is introduced in \cite{sultani2018ranking} to harness the high-level class labels to directly learn the anomaly score for each video segment (\ie, a small number of consecutive video frames). Its key objective is to guarantee that the maximum anomaly score for the segments in a video that contains anomalies somewhere is greater than the counterparts in a normal video. To achieve this, each video is treated as a bag of instances in MIL, the videos that contains anomalies are treated as positive bags, and the normal videos are treated as negative bags. Each video segment is an instance in the bag. The ordering information of the anomaly scores is enforced as a relative pairwise ranking order via the hinge loss function. The overall objective function is defined as
\begin{multline}
    \argmin_{\Theta} \sum_{\mathcal{B}_p, \mathcal{B}_n \in \mathcal{X} } \max\{0, 1-\max_{\mathbf{x} \in \mathcal{B}_p} \tau(\mathbf{x};\Theta) + \max_{\mathbf{x} \in \mathcal{B}_n} \tau(\mathbf{x};\Theta)\}\\
    + \lambda_1 \sum_{i=1}^{|\mathcal{B}_p|} \big(\tau(\mathbf{x}_i;\Theta)-\tau(\mathbf{x}_{i+1};\Theta)\big)^2 + \lambda_2 \sum_{\mathbf{x} \in \mathcal{B}_{p}} \tau(\mathbf{x};\Theta),
\end{multline}
where $\mathbf{x}$ is a video segment, $\mathcal{B}$ contains a bag of video segments, and $\mathcal{B}_p$ and $\mathcal{B}_n$ respectively represents positive and negative bags. The first term is to guarantee the relative anomaly score order, \ie, the anomaly score of the most abnormal video segment in the positive instance bag is greater than that in the negative instance bag. The last two terms are extra optimization constraints, in which the former enforces score smoothness between consecutive video segments while the latter enforces anomaly sparsity, \ie, each video contains only a few abnormal segments.

\textit{Advantages}. The advantages of deep ranking model-base methods are as follows. (i) The anomaly scores can be optimized directly with adapted loss functions. (ii) They are generally free from the definitions of anomalies by imposing a weak assumption of the ordinal order between anomaly and normal instances. (iii) This approach may build upon well-established ranking techniques and theories from areas like learning to rank \cite{liu2009ranking,wang2018learningtorank,liu2018learningtorank}. 
% Although they address very different problems (\eg, the queries in learning to rank do not exist in anomaly detection), they may share some key objectives (\eg, the target of sorting data instances).

\textit{Disadvantages}. Their disadvantages are as follows. (i) At least some form of labeled anomalies are required in these methods, which may not be applicable to applications where such labeled anomalies are not available. The method in \cite{pang2020ranking} is fully unsupervised and obtains some promising performance but there is still a large gap compared to semi-supervised methods. (ii) Since the models are exclusively fitted to detect the few labeled anomalies, they may not be able to generalize to unseen anomalies that exhibit different abnormal features to the labeled anomalies.

\textit{Challenges Targeted}: Using weak supervision such as pseudo labels or noisy class labels provide some important knowledge of suspicious anomalies, enabling the learning of more expressive low-dimensional representation space and better detection accuracy (\textbf{CH1, CH2}). The MIL scheme \cite{sultani2018ranking} and the pairwise relation prediction \cite{pang2019ranking} provide an easy way to incorporate coarse-grained/limited anomaly labels to detection model learning (\textbf{CH3}). More importantly, the end-to-end anomaly score learning offers straightforward anomaly explanation by backpropagating the activation weights or the gradient of anomaly scores to locate the features that are responsible for large anomaly scores \cite{pang2020ranking} (\textbf{CH6}). In addition, the methods in \cite{pang2019ranking,pang2020ranking} also work well in data with anomaly contamination or noisy labels (\textbf{CH4}).

\subsection{Prior-driven Models}\label{subsec:prior}
This approach uses a prior distribution to encode and drive the anomaly score learning. Since the anomaly scores are learned in an end-to-end manner, the prior may be imposed on either the internal module or the learning output (\ie, anomaly scores) of the score learning function $\tau$.

\textit{Assumptions}. The imposed prior captures the underlying (ab)normality of the dataset.

The incorporation of the prior into the internal anomaly scoring function is exemplified by a recent study on the Bayesian inverse reinforcement learning (IRL)-based method \cite{oh2019prior}. The key intuition is that given an agent that takes a set of sequential data as input, the agent's normal behavior can be understood by its latent reward function, and thus a test sequence is identified as anomaly if the agent assigns a low reward to the sequence. IRL approaches \cite{ng2000irl} are used to infer the reward function. To learn the reward function more efficiently, a sample-based IRL approach is used. Specifically, the IRL problem is formulated as the below posterior optimization problem 
\begin{equation}\label{eqn:irl}
    \max_{\Theta} \mathbb{E}_{\mathbf{s}\sim\mathcal{S}}\big[\log p(\mathbf{s}|\Theta) + \log p(\Theta)\big],
\end{equation}
where $p(\mathbf{s}|\Theta)=\frac{1}{Z}\exp\big(\sum_{(o,a)\in\mathbf{s}}\tau_{\Theta}(o,a)\big)$, $\tau_{\Theta}(o,a)$ is a latent reward function parameterized by $\Theta$, $(o,a)$ is a pair of state and action in the sequence $\mathbf{s}$, $Z$ represents the partition function which is the integral of $\exp\big(\sum_{(o,a)\in\mathbf{s}}\tau_{\Theta}(o,a)\big)$ over all the sequences consistent with the underlying Markov decision process dynamics, $p(\Theta)$ is a prior distribution of $\Theta$, and $\mathcal{S}$ is a set of observed sequences. Since the inverse of the reward yielded by $\tau$ is used as the anomaly score, maximizing Eq. (\ref{eqn:irl}) is equivalent to directly learning the anomaly scores.

At the training stage, a Gaussian prior distribution over the weight parameters of the reward function learning network is assumed, \ie, $\Theta \sim \mathcal{N}(0, \sigma^2)$. The partition function $Z$ is estimated using a set of sequences generated by a sample-generating policy $\pi$, 
\begin{equation}
    Z = \mathbb{E}_{\mathbf{s} \sim \pi} \big[ \sum_{(o,a) \in \mathbf{s}} \tau_{\Theta}(o,a) \big].
\end{equation}
The policy $\pi$ is also represented as a neural network. $\tau$ and $\pi$ are alternatively optimized, \ie, to optimize the reward function $\tau$ with a fixed policy $\pi$ and to optimize $\pi$ with the updated reward function $\tau$. Note that $\tau$ is instantiated with a bootstrap neural network with multiple output heads in \cite{oh2019prior}; Eq. (\ref{eqn:irl}) presents a simplified $\tau$ for brevity.

The idea of enforcing a prior on the anomaly scores is explored in \cite{pang2019devnet}. Motivated by the extensive empirical results in \cite{kriegel2011interpreting} that show the anomaly scores in a variety of real-world data sets fits Gaussian distribution very well, the work uses a Gaussian prior to encode the anomaly scores and enable the direct optimization of the scores. That is, it is assumed that the anomaly scores of normal instances are clustered together while that of anomalies deviate far away from this cluster. The prior is leveraged to define a loss function, called deviation loss, which is built upon the well-known contrastive loss \cite{hadsell2006contrastloss}. 

\begin{align}\label{eqn:deviationloss}
    L_{\mathit{dev}} = (1-y_{\mathbf{x}})|\mathit{dev}(\mathbf{x})| + y_{\mathbf{x}} \max\big\{0, m - \mathit{dev}(\mathbf{x})\big\} \quad  \text{and} \quad  \mathit{dev}(\mathbf{x}) = \frac{\tau(\mathbf{x};\Theta) - \mu_{b}}{\sigma_{b}},
\end{align}
where $\mu_{b}$ and $\sigma_{b}$ are respectively the estimated mean and standard deviation of the prior $\mathcal{N}(\mu, \sigma)$, $y_{\mathbf{x}}=1$ if $\mathbf{x}$ is an anomaly and $y_{\mathbf{x}}=0$ if $\mathbf{x}$ is a normal object, and $m$ is equivalent to a Z-Score confidence interval parameter. $\mu_b$ and $\sigma_b$ are estimated using a set of values $\{r_1, r_2, \cdots, r_l\}$ drawn from $\mathcal{N}(\mu, \sigma)$ for each batch of instances to learn robust representations of normality and abnormality.

The detection model is driven by the deviation loss to push the anomaly scores of normal instances as close as possible to $\mu$ while guaranteeing at least $m$ standard deviations between $\mu$ and the anomaly scores of anomalies. When $\mathbf{x}$ is an anomaly and it has a negative $\mathit{dev}(\mathbf{x})$, the loss would be particularly large, resulting in large \textit{positive deviations} for all anomalies. As a result, the deviation loss is equivalent to enforcing a statistically significant deviation of the anomaly score of the anomalies from that of normal instances in the upper tail. Further, this Gaussian prior-driven loss also results in well interpretable anomaly scores, \ie, given any anomaly score $\tau(\mathbf{x})$, we can use the Z-score confidence interval $\mu \pm z_{p}\sigma $ to explain the abnormality of the instance $\mathbf{x}$. This is an important and very practical property that existing methods do not have.

\textit{Advantages}. The advantages of prior-driven models are as follows. (i) The anomaly scores can be directly optimized w.r.t. a given prior. (ii) It provides a flexible framework for incorporating different prior distributions into the anomaly score learning. Different Bayesian deep learning techniques \cite{wang2016bayesiandl} may be adapted for anomaly detection. (iii) The prior can also result in more interpretable anomaly scores than the other methods.

\textit{Disadvantages}. Their disadvantages are as follows. (i) It is difficult, if not impossible, to design a universally effective prior for different anomaly detection application scenarios. (ii) The models may work less effectively if the prior does not fit the underlying distribution well.

\textit{Challenges Targeted}: The prior empowers the models to learn informed low-dimensional representations of different complex data such as high-dimensional data and sequential data (\textbf{CH1 \& CH2}). By imposing a prior over anomaly scores, the deviation network method \cite{pang2019devnet} shows promising performance in leveraging a limited amount of labeled anomaly data to enhance the representations of normality and abnormality, substantially boosting the detection recall (\textbf{CH1 \& CH3}). The detection models here are driven by a prior distribution w.r.t. anomaly scoring function and work well in data with anomaly contamination in the training data (\textbf{CH4}).

\subsection{Softmax Likelihood Models}\label{subsec:mle}

This approach aims at learning anomaly scores by maximizing the likelihood of events in the training data. Since anomaly and normal instances respectively correspond to rare and frequent patterns, from the probabilistic perspective, normal instances are presumed to be high-probability events whereas anomalies are prone to be low-probability events. Therefore, the negative of the event likelihood can be naturally defined as anomaly score. Softmax likelihood models are shown effective and efficient in achieving this goal via tools like noise contrastive estimation (NCE) \cite{gutmann2010nce}. 

\textit{Assumptions}. Anomalies and normal instances are respectively low- and high-probability events.

The idea of learning anomaly scores by directly modeling the event likelihood is introduced in \cite{chen2016mle}. Particularly, the problem is framed as

\begin{equation}\label{eqn:mle}
    \Theta^*=\argmax_{\Theta} \sum_{\mathbf{x} \in \mathcal{X}} \log p(\mathbf{x}; \Theta),
\end{equation}
where $p(\mathbf{x}; \Theta)$ is the probability of the instance $\mathbf{x}$ (\ie, an event in the event space) with the parameters $\Theta$ to be learned. To easy the optimization, $p(\mathbf{x}; \Theta)$ is modeled with a softmax function:
\begin{equation}\label{eqn:softmax}
    p(\mathbf{x}; \Theta) = \frac{\exp \big(\tau(\mathbf{x}; \Theta)\big)}{\sum_{\mathbf{x} \in \mathcal{X}} \exp \big(\tau(\mathbf{x}; \Theta)\big)},
\end{equation}
where $\tau(\mathbf{x}; \Theta)$ is an anomaly scoring function designed to capture pairwise feature interactions:
\begin{equation}\label{eqn:mlescore}
    \tau(\mathbf{x}; \Theta) = \sum_{i,j \in \{1, 2, \cdots, K\}} w_{ij}\mathbf{z}_{i}\mathbf{z}_{j},
\end{equation}
where $\mathbf{z}_{i}$ is a low-dimensional embedding of the $i$th feature value of $\mathbf{x}$ in the representation space $\mathcal{Z}$, $w_{ij}$ is the weight added to the interaction and is a trainable parameter. Since $\sum_{\mathbf{x} \in \mathcal{X}} \exp \big(\tau(\mathbf{x}; \Theta)\big)$ is a normalization term, learning the likelihood function $p$ is equivalent to directly optimizing the anomaly scoring function $\tau$. The computation of this explicit normalization term is prohibitively costly, the well-established NCE is used in \cite{chen2016mle} to learn the following approximated likelihood
\begin{equation}
   \log p(d=1|\mathbf{x};\Theta) + \log \sum_{j=1}^{k} p(d=0|\mathbf{x}^{\prime}_{j};\Theta),
\end{equation}
where $p(d=1|\mathbf{x};\Theta) = \frac{\exp\big(\tau(\mathbf{x}; \Theta)\big)}{\exp\big(\tau(\mathbf{x}; \Theta)\big) + k Q(\mathbf{x}^{\prime})}$ and $p(d=0|\mathbf{x}^{\prime};\Theta) = \frac{k Q(\mathbf{x}^{\prime})}{\exp\big(\tau(\mathbf{x}; \Theta)\big) + k Q(\mathbf{x}^{\prime})}$; for each instance $\mathbf{x}$, $k$ noise samples $\mathbf{x}^{\prime}_{1,\cdots,k} \sim Q$ are generated from some synthetic known `noise' distribution $Q$. In \cite{chen2016mle}, a context-dependen method is used to generate the $k$ negative samples by univariate extrapolation of the observed instance $\mathbf{x}$. 

The method is primarily designed to detect anomalies in categorical data \cite{chen2016mle}. Motivated by this application, a similar objective function is adapted to detect abnormal events in heterogeneous attributed bipartite graphs \cite{fan2018mle}. The problem in \cite{fan2018mle} is to detect anomalous paths that span both partitions of the bipartite graph. Therefore, $\mathbf{x}$ in Eq. (\ref{eqn:mlescore}) is a graph path containing a set of heterogeneous graph nodes, with $\mathbf{z}_i$ and $\mathbf{z}_j$ be the representations of every pair of the nodes in the path. To map attributed nodes into the representation space $\mathcal{Z}$, multilayer perceptron networks and autoencoders are respectively applied to the node features and the graph topology.

\textit{Advantages}. The advantages of softmax model-based methods are as follows. (i) Different types of interactions can be incorporated into the anomaly score learning process. (ii) The anomaly scores are faithfully optimized w.r.t. the specific abnormal interactions we aim to capture.

\textit{Disadvantages}. Their disadvantages are as follows. (i) The computation of the interactions can be very costly when the number of features/elements in each data instance is large, \ie, we have $O(D^{n})$ time complexity per instance for $n$-th order interactions of $D$ features/elements. (ii) The anomaly score learning is heavily dependent on the quality of the generation of negative samples.

\textit{Challenges Targeted}: The formulation in this category of methods provides a promising way to learn low-dimensional representations of datasets with heterogeneous data sources (\textbf{CH2 \& CH5}). The learned representations often capture more normality/abnormality information from different data sources and thus enable better detection than traditional methods (\textbf{CH1}).
% The methods learn event/instance likelihood models to detect low-probability events as anomalies, and thus, they are not bounded by specific anomaly measures  (\textbf{CH4}).

\subsection{End-to-end One-class Classification}\label{subsec:oneclasse2e}
This category of methods aims to train a one-class classifier that learns to discriminate whether a given instance is normal or not in an end-to-end manner. Different from the methods in Section \ref{subsubsec:oneclass}, this approach does not rely on any existing one-class classification measures such as one-class SVM or SVDD. This approach emerges mainly due to the marriage of GANs and the concept of one-class classification, \ie, adversarially learned one-class classification. The key idea is to learn a one-class discriminator of the normal instances so that it well discriminates those instances from adversarially generated pseudo anomalies. This approach is also very different from the GAN-based methods in Section \ref{subsubsec:generative} due to two key differences. First, the GAN-based methods aim to learn a generative distribution to maximally approximate the real data distribution, achieving a generative model that well captures the normality of the training normal instances; while the methods in this section aim to optimize a discriminative model to separate normal instances from adversarially generated fringe instances. Second, the GAN-based methods define the anomaly scores based on the residual between the real instances and the corresponding generated instances, whereas the methods here directly use the discriminator to classify anomalies, \ie, the discriminator $D$ acts as $\tau$ in Eq. (\ref{eqn:scorelearning1}). This section is separated from Sections \ref{subsubsec:generative} and \ref{subsubsec:oneclass} to highlight the above differences.

\textit{Assumptions}. 
% Two basic assumptions of this approach are as follows.
(i) Data instances that are approximated to anomalies can be effectively synthesized. (ii) All normal instances can be summarized by a discriminative one-class model.

The idea of adversarially learned one-class (ALOCC) classification is first studied in \cite{sabokrou2018oneclasse2e}. The key idea is to train two deep networks, with one network trained as the one-class model to separate normal instances from anomalies while the other network trained to enhance the normal instances and generate distorted outliers. The two networks are instantiated and optimized through the GANs approach. The one-class model is built upon the discriminator network and the generator network is based on a denoising AE \cite{vincent2010denoisingae}. The objective of the AE-empower GAN is defined as

\begin{equation}\label{eqn:alocc}
    \min_{\mathit{AE}} \max_{D} V(D, G) = \mathbb{E}_{\mathbf{x} \sim p_{\mathcal{X}} } \big[\log D(\mathbf{x})\big] +  \mathbb{E}_{\hat{\mathbf{x}} \sim p_{\hat{\mathcal{X}}} } \bigg[\log\Big(1- D\big(\mathit{AE}(\hat{\mathbf{x}})\big)\Big)\bigg],
\end{equation}
where $p_{\hat{\mathcal{X}}}$ denotes a data distribution of $\mathcal{X}$ corrupted by a Gaussian noise, \ie, $\hat{\xvec} = \xvec + \mathbf{n}$ with $\mathbf{n} \sim \mathcal{N}(0, \sigma^2\mathbf{I})$. This objective is jointly optimized with the following data construction error in $\mathit{AE}$.

\begin{equation}\label{eqn:dae}
    \ell_{\mathit{ae}}= \|\xvec - \mathit{AE}(\hat{\xvec})\|^2.
\end{equation}

The intuition in Eq. (\ref{eqn:alocc}) is that $\mathit{AE}$ can well reconstruct (and even enhance) normal instances, but it can be confused by input outliers and consequently generates distorted outliers. Through the minimax optimization, the discriminator $D$ learns to better discriminate normal instances from the outliers than using the original data instances. Thus, $D\big(\mathit{AE}(\hat{\mathbf{x}})\big)$ can be directly used to detect anomalies. In \cite{sabokrou2018oneclasse2e} the outliers are randomly drawn from some classes other than the classes where the normal instances come from. 

% The OCGAN model introduced in \cite{perera2019oneclasse2e} improves ALOCC by adding an extra discriminator, called latent discriminator, which is incorporated to learn a latent feature space that exclusively represents the the given normal training data instances. The objective of this discriminator is as follows

% \begin{multline}
%     \max_{D} \mathbb{E}_{\zvec \sim \mathbb{U}(-1, 1) }[\log D(\zvec)] + \mathbb{E}_{\hat{\xvec} \sim p_{\mathcal{X}}} \big[\log\big(1 - D(E(\hat{\xvec}))\big) \big] ,
% \end{multline}

However, obtaining the reference outliers beyond the given training data as in \cite{sabokrou2018oneclasse2e} may be unavailable in many domains. Instead of taking random outliers from other datasets, we can generate fringe data instances based on the given training data and use them as negative reference instances to enable the training of the one-class discriminator. This idea is explored in \cite{zheng2019oneclasse2e,ngo2019oneclasse2e}. One-class adversarial networks (OCAN) is introduced in \cite{zheng2019oneclasse2e} to leverage the idea of bad GANs \cite{dai2017badgan} to generate fringe instances based on the distribution of the normal training data. Unlike conventional generators in GANs, the generator network in bad GANs is trained to generate data instances that are complementary, rather than matching, to the training data. The objective of the complement generator is as follows

\begin{equation}\label{eqn:complementgenerator}
    \min_{G} -\mathcal{H}(p_{\mathcal{Z}}) + \mathbb{E}_{\hat{\zvec} \sim p_{\mathcal{Z}}} \log p_{\mathcal{X}}(\hat{\zvec})\mathbb{I}[p_{\mathcal{X}}(\hat{\zvec})> \epsilon]+  \| \mathbb{E}_{\hat{\zvec} \sim p_{\mathcal{Z}}} h(\hat{\zvec}) -  \mathbb{E}_{\zvec \sim p_{\mathcal{X}}} h(\zvec)\|_2,
\end{equation}
where $\mathcal{H}(\cdot)$ is the entropy, $\mathbb{I}[\cdot]$ is an indicator function, $\epsilon$ is a threshold hyperparameter, and $h$ is a feature mapping derived from an intermediate layer of the discriminator. The first two terms are devised to generate low-density instances in the original feature space. However, it is computationally infeasible to obtain the probability distribution of the training data. Instead the density estimation $p_{\mathcal{X}}(\hat{\zvec})$ is approximated by the discriminator of a regular GAN. The last term is the widely-used feature matching loss that helps better generate data instances within the original data space. The objective of the discriminator in OCAN is enhanced with an extra conditional entropy term to enable the detection with high confidence:
\begin{equation}
    \max_{D} \mathbb{E}_{\xvec \sim p_{\mathcal{X}}} \big[\log D(\zvec)\big] + \mathbb{E}_{\hat{\zvec} \sim p_{\mathcal{Z}}} \big[\log\big(1 - D(\hat{\zvec})\big) \big] + \mathbb{E}_{\xvec \sim p_{\mathcal{X}}}\big[D(\xvec)\log D(\xvec)\big],
\end{equation}

In \cite{ngo2019oneclasse2e}, Fence GAN is introduced with the objective to generate data instances tightly lying at the boundary of the distribution of the training data. This is achieved by introducing two loss functions into the generator that enforce the generated instances to be evenly distributed along a sphere boundary of the training data. Formally, the objective of the generator is defined as 
\begin{equation}
    \min_{G} \mathbb{E}_{\zvec \sim p_{\mathcal{Z}}} \Big[ \log \big[ \big| \alpha - D\big(G(\zvec)\big) \big| \big] \Big] + \beta \frac{1}{\mathbb{E}_{\zvec \sim p_{\mathcal{Z}}} \| G(\zvec) -\boldsymbol{\mu} \|_2},
\end{equation}
where $\alpha \in (0, 1)$ is a hyperparameter used as a discrimination reference score for the generator to generate the boundary instances and $\boldsymbol{\mu}$ is the center of the generated data instances. The first term is called encirclement loss that enforces the generated instances to have the same discrimination score, ideally resulting in instances tightly enclosing the training data. The second term is called dispersion loss that enforces the generated instances to evenly cover the whole boundary.

% \begin{multline}
%     \max_{D} \frac{1}{N}\sum_{i=1}^{N} \bigg[- \log \big( D(\xvec_i) \big) - \gamma \log \Big( 1 - D\big(G(\zvec_i)\big) \Big)   \bigg],
% \end{multline}

There have been some other methods introduced to effectively generate the reference instances. For example, uniformly distributed instances can be generated to enforce the normal instances to be distributed uniformly across the latent space \cite{perera2019oneclasse2e}; an ensemble of generators is used in \cite{liu2019oneclasse2e}, with each generator synthesizing boundary instances for one specific cluster of normal instances.

\textit{Advantages}. The advantages of this category of methods is as follows. (i) Its anomaly classification model is adversarially optimized in an end-to-end fashion. (ii) It can be developed and supported by the affluent techniques and theories of adversarial learning and one-class classification.

\textit{Disadvantages}. Their disadvantages are as follows. (i) It is difficult to guarantee that the generated reference instances well resemble the unknown anomalies. (ii) The instability of GANs may lead to generated instances with diverse quality and consequently unstable anomaly classification performance. This issue is recently studied in \cite{zaheer2020oneclasse2e}, which shows that the performance of this type of anomaly detectors can fluctuate drastically in different training steps. (iii) Its applications are limited to semi-supervised anomaly detection scenarios.

\textit{Challenges Targeted}: The adversially learned one-class classifiers learn to generate realistic fringe/boundary instances, enabling the learning of expressive low-dimensional normality representations (\textbf{CH1 \& CH2}). 

% The methods model anomaly detection as a pseudo anomaly classification problem and thus they are anomaly-definition-agnostic (\textbf{CH4}).

\section{Algorithms and Datasets}\label{sec:algorithm}

\subsection{Representative Algorithms}

To gain a more in-depth understanding of methods in this area, in Table \ref{tab:method} we summarize some key characteristics of representative algorithms from each category of methods. Since these methods are evaluated on diverse datasets, it is difficult to have an universal meta-analysis of their empirical performance. Instead, some main observations w.r.t. the model design are summarized as follows: (i) most methods operate in an unsupervised or semi-supervised mode; (ii) deep learning tricks like data augmentation, dropout and pre-training are under-explored; (iii) the network architecture used is not that deep, with a majority of the methods having no more than five network layers; (iv) (leaky) ReLU is the most popular activation function; and (v) diverse backbone networks can be used to handle different types of input data. The source code of most of these algorithms is publicly accessible. We summarize those source codes in Table A1 in Appendix A to facilitate the access.
%  in Appendix \ref{sec:sourcecode}

\begin{table*}[htbp]
\centering
\caption{Key Characteristics of 30 Representative Algorithms. DA, DP, PT and Archit. are short for data augmentation, dropout, pre-training and architecture, respectively. \# layers account for all layers except the input layer. lReLU represents leaky ReLU.}
  \scalebox{0.65}{
    \begin{tabular}{cccccccc@{}c@{}c@{}c@{}c}
    \hline\hline
    % \multicolumn{5}{||c||}{\textbf{Data Characteristic}} & \multicolumn{5}{|c||}{\textbf{AUC-ROC Performance}}               & \multicolumn{5}{c||}{\textbf{AUC-PR Performance}} \\\hline
    \textbf{Method} & \textbf{Ref.} & \textbf{Sup.} & \textbf{Objective}  & \textbf{DA} & \textbf{DP} & \textbf{PT} & \textbf{Archit.} & \textbf{Activation} & \textbf{\# layers} & \textbf{Loss} & \textbf{Data }\\\hline
    OADA & \cite{ionescu2019featureextraction} (\ref{sec:featureextraction}) & Semi & Reconstruction &  Yes & No& No & AE & ReLU & 3 & MSE & Video \\
    Replicator &\cite{hawkins2002autoencoder} (\ref{subsubsec:ae}) & Unsup. & Reconstruction & No & No& No & AE & Tanh& 2 & MSE  & Tabular\\
    RandNet &\cite{chen2017ensembleae} (\ref{subsubsec:ae}) & Unsup. & Reconstruction & No & Yes& Yes  & AE & ReLU & 3 & MSE  & Tabular\\
    RDA &\cite{zhou2017autoencoder} (\ref{subsubsec:ae}) & Semi & Reconstruction & No & No &No& AE  & Sigmoid&2&  MSE   & Tabular\\
    UODA &\cite{lu2017sequenceae} (\ref{subsubsec:ae}) & Semi & Reconstruction & No &  No & Yes & AE \& RNN & Sigmoid & 4 & MSE  & Sequence\\
    AnoGAN & \cite{schlegl2017generative} (\ref{subsubsec:generative}) & Semi & Generative & No & No &No & Conv. & ReLU &  4 & MAE & Image\\
    EBGAN & \cite{zenati2018generative} (\ref{subsubsec:generative}) & Semi & Generative  & No & No &No &  Conv. \& MLP& ReLU/lReLU & 3-4  & GAN & Image \& Tabular\\
    FFP & \cite{liu2018predictabilitycvpr} (\ref{subsec:selfprediction}) & Semi & Predictive & Yes & No & Yes & Conv. & ReLU & 10 & MAE/MSE & Video \\
    LSA & \cite{abati2019predictabilitycvpr} (\ref{subsec:selfprediction}) & Semi & Predictive & No & No &No &  Conv. &lReLU & 4-7& MSE \& KL & video\\
    GT &  \cite{golan2018featuretrasnformation} (\ref{subsubsec:consistency}) & Semi & Classification & Yes & Yes & No & Conv. & ReLU & 10-16 & CE&Image \\
    E$^3$Outlier & \cite{wang2019featuretransformation} (\ref{subsubsec:consistency}) & Semi & Classification & Yes & Yes & No& Conv. & ReLU & 10 & CE&Image \\
    REPEN &\cite{pang2018repen} (\ref{subsubsec:distancemeasure}) & Unsup. & Distance & No & No& No & MLP & ReLU & 1 &  Hinge  & Tabular\\
    RDP & \cite{wang2020distance} (\ref{subsubsec:distancemeasure}) & Unsup. & Distance & No & No & No& MLP & lReLU & 1 & MSE  & Tabular\\
    AE-1SVM & \cite{nguyen2018oneclasssvm} (\ref{subsubsec:oneclass}) & Unsup. & One-class & No & No & No& AE \& Conv. & Sigmoid & 2-5 & Hinge  & Tabular \& image\\
    % OC-NN & \cite{chalapathy2018oneclasssvm} (\ref{subsubsec:oneclass}) & Unsup. & One-class & No& No& Yes & MLP & Linear  & 1-2 & Hinge  & Tabular \\
    DeepOC & \cite{wu2019oneclasssvm}(\ref{subsubsec:oneclass}) &Semi  & One-class & No & No &No& 3D Conv. & ReLU & 5 & Hinge  & Video\\
    Deep SVDD & \cite{ruff2018deepsvdd} (\ref{subsubsec:oneclass}) & Semi & One-class & No & No & Yes & Conv. & lReLU & 3-4 & Hinge  & Image\\
    Deep SAD & \cite{ruff2020deepsvdd} (\ref{subsubsec:oneclass}) & Semi & One-class & No & No & Yes & Conv. \& MLP & lReLU  & 3-4 & Hinge  & Image \& Tabular\\ 
    DEC & \cite{xie2016clustering} (\ref{subsubsec:clustering})& Unsup. & Clustering &  No & Yes & Yes & MLP & ReLU & 4 & KL & Image \& Tabular\\
    DAGMM & \cite{zong2018autoencoder} (\ref{subsubsec:clustering}) & Unsup. & Clustering &  No & Yes & No & AE \& MLP & Tanh& 4-6 & Likelihood  & Tabular \\
    SDOR & \cite{pang2020ranking} (\ref{subsec:ranking}) & Unsup. & Anomaly scores & No & No & Yes & ResNet \& MLP & ReLU & 50 + 2 & MAE  & Video\\
    PReNet & \cite{pang2019ranking} (\ref{subsec:ranking}) & Weak & Anomaly scores & Yes & No & No& MLP & ReLU & 2-4& MAE & Tabular\\
    MIL & \cite{sultani2018ranking} (\ref{subsec:ranking}) & Weak & Anomaly scores & No & Yes & Yes & 3DConv. \& MLP & ReLU & 18/34 + 3 & Hinge  & Video\\ 
    PUP & \cite{oh2019prior} (\ref{subsec:prior}) & Unsup. & Anomaly scores &  No& No & No & MLP & ReLU & 3 & Likelihood  & Sequence\\
    DevNet &\cite{pang2019devnet} (\ref{subsec:prior}) & Weak & Anomaly scores & No & No& No & MLP & ReLU & 2-4 & Deviation  & Tabular\\
    APE & \cite{chen2016mle} (\ref{subsec:mle}) & Unsup. & Anomaly scores & No & No & No& MLP & Sigmoid & 3 & Softmax  &Tabular\\
    AEHE & \cite{fan2018mle} (\ref{subsec:mle}) & Unsup. & Anomaly scores & No & No & No& AE \& MLP & ReLU & 4 & Softmax  & Graph\\
    ALOCC & \cite{sabokrou2018oneclasse2e}  (\ref{subsec:oneclasse2e}) & Semi & Anomaly scores & Yes & No & No& AE \& CNN & lReLU & 5 & GANs & Image\\
    OCAN & \cite{zheng2019oneclasse2e} (\ref{subsec:oneclasse2e}) & Semi & Anomaly scores & No & No& Yes & LSTM-AE \& MLP & ReLU & 4 & GANs  & Sequence\\
    Fence GAN & \cite{ngo2019oneclasse2e} (\ref{subsec:oneclasse2e}) & Semi & Anomaly scores & No & Yes & No & Conv. \& MLP & lReLU/Sigmoid & 4-5 & GANs & Image \& Tabular\\
    OCGAN  & \cite{perera2019oneclasse2e}  (\ref{subsec:oneclasse2e}) & Semi & Anomaly scores & No & No & No & Conv. & ReLU/Tanh & 3 & GANs & Image \\ 
    \hline\hline
    \end{tabular}
    }
\label{tab:method}
\end{table*}

\subsection{Datasets with Real Anomalies}

One main obstacle to the development of anomaly detection is the lack of real-world datasets with real anomalies. Many studies (\eg, \cite{zhou2017autoencoder,zenati2018generative,akcay2018generative,golan2018featuretrasnformation,wang2019featuretransformation,ruff2018deepsvdd,ngo2019oneclasse2e}) evaluate the performance of their presented methods on datasets converted from popular classification data for this reason. This way may fail to reflect the performance of the methods in real-world anomaly detection applications. 
% as the characteristics of anomalies in the converted data can be different from the real anomalies in practice
We summarize a collection of 21 publicly available real-world datasets with real anomalies in Table \ref{tab:application} to promote the performance evaluation on these datasets. The datasets cover a wide range of popular application domains presented in a variety of data types. Only large-scale and/or high-dimensional complex datasets are included here to provide challenging testbeds for deep anomaly detection. In addition, a continuously updated collection of widely-used anomaly detection datasets (including some pre-processed datasets from Table \ref{tab:application}) is made available at \url{https://git.io/JTs93}.

\begin{table*}[htbp]
\centering
\caption{21 Publicly Accessible Real-world Datasets with Real Anomalies.}
  \scalebox{0.65}{
    \begin{tabular}{cc@{}c@{}cc@{}ccc}
    \hline\hline
    % \multicolumn{5}{||c||}{\textbf{Data Characteristic}} & \multicolumn{5}{|c||}{\textbf{AUC-ROC Performance}}               & \multicolumn{5}{c||}{\textbf{AUC-PR Performance}} \\\hline
    % \footnote{http://kdd.ics.uci.edu/databases/kddcup99/kddcup99.html}
    %\footnote{https://archive.ics.uci.edu/ml/datasets/internet+advertisements}
    % \footnote{http://archive.ics.uci.edu/ml/datasets/Thyroid+Disease}
    %\footnote{http://archive.ics.uci.edu/ml/datasets/Arrhythmia}
    % \footnote{https://www.kaggle.com/c/kdd-cup-2014-predicting-excitement-at-donors-choose}
    % \footnote{https://biendata.com/competition/kddcup2015/}
    \textbf{Domain} & \textbf{Data} & \textbf{Size} & \textbf{Dimension} &  \textbf{Anomaly (\%)} &  \textbf{Type} & \textbf{Reference} \\\hline
    Intrusion detection & KDD Cup 99 \cite{bache2013uci} & 4,091-567,497 & 41 & 0.30\%-7.70\% & Tabular &  \cite{hawkins2002autoencoder,zong2018autoencoder,nguyen2018oneclasssvm,ngo2019oneclasse2e} \\
    Intrusion detection & UNSW-NB15 \cite{moustafa2015nb15} & 257,673 & 49 & $\leq$9.71\% &  Streaming & \cite{pang2019ranking,pang2019devnet}\\
    Excitement prediction & KDD Cup 14 & 619,326 & 10  & 6.00\% &  Tabular & \cite{pang2019ranking,pang2019devnet}\\
    Dropout prediction & KDD Cup 15 & 35,091  &27 &  0.10\%-0.40\% &  Sequence &  \cite{lu2017sequenceae} \\
    Malicious URLs detection & URL \cite{ma2009url} & 2.4m & 3.2m &  33.04\% &  Streaming & \cite{pang2018repen}\\
    Spam detection & Webspam \cite{webb2006webspam} & 350,000 & 16.6m &  39.61\% &  Tabular/text &\cite{pang2018repen}\\
    Fraud detection & Credit-card-fraud \cite{dal2017creditcardfraud} & 284,807 & 30 &  0.17\% &  Streaming & \cite{zheng2019oneclasse2e,pang2019ranking,pang2019devnet}\\
    Vandal detection & UMDWikipedia \cite{kumar2015wikipedia} & 34,210 & N/A &  50.00\% &  Sequence &  \cite{zheng2019oneclasse2e} \\
    Mutant activity detection & p53 Mutants \cite{bache2013uci} & 16,772 & 5,408 & 0.48\% &   Tabular &\cite{pang2018repen}\\
    Internet ads detection & AD \cite{bache2013uci} & 3,279 & 1,555& 14.00\% &   Tabular &\cite{pang2018repen}\\
    Disease detection & Thyroid \cite{bache2013uci} & 7,200 & 21 & 7.40\% &   Tabular &\cite{pang2019ranking,pang2019devnet,zong2018autoencoder,ruff2020deepsvdd}\\
    Disease detection & Arrhythmia \cite{bache2013uci} & 452 & 279 & 14.60\%&   Tabular &\cite{pang2015lesinn,zong2018autoencoder,ruff2020deepsvdd}\\
    Defect detection & MVTec AD & 5,354 & N/A & 35.26\% & Image & \cite{bergmann2019mvtec}\\
    Video surveillance & UCSD Ped 1 \cite{li2013ucsd} &  14,000 frames & N/A& 28.6\% &   Video &\cite{pang2020ranking,wu2019oneclasssvm}\\
    Video surveillance & UCSD Ped 2 \cite{li2013ucsd} &  4,560 frames & N/A& 35.9\% &   Video &\cite{pang2020ranking,wu2019oneclasssvm}\\
    Video surveillance & UMN \cite{umndata} &   7,739 frames & N/A& 15.5\%- 18.1\% &   Video &\cite{pang2020ranking}\\
    Video surveillance & Avenue \cite{lu2013avenue} & 30,652 frames & N/A & 12.46\% &   Video & \cite{wu2019oneclasssvm}\\
    Video surveillance &  ShanghaiTech Campus  & 317,398 frames & N/A& 5.38\% &   Video & \cite{liu2018predictabilitycvpr}  \\
    Video surveillance &  UCF-Crime  &  1,900 videos (13.8m frames) & N/A& 13 crimes &  Video &  \cite{sultani2018ranking} \\
    System log analysis & HDFS Log \cite{xu2009hdfs} & 11.2m & N/A & 2.90\% &  Sequence & \cite{du2017deeplog}\\
    System log analysis & OpenStack log & 1.3m & N/A & 7.00\% &  Sequence & \cite{du2017deeplog}\\
    \hline\hline
    \end{tabular}
    }
\label{tab:application}
\end{table*}

\section{Conclusions and Future Opportunities}

In this work we review 12 diverse modeling perspectives on harnessing deep learning techniques for anomaly detection. We also discuss how these methods address some notorious anomaly detection challenges to demonstrate the importance of deep anomaly detection. Through such a review, we identify some exciting opportunities as follows.

\subsection{Exploring Anomaly-supervisory Signals}

Informative supervisory signals are the key for deep anomaly detection to learn accurate anomaly scores or expressive representations of normality/abnormality.
% and reduce false positives. 
%However, anomalies are often unknown and have diverse causes. Thus, it is difficult, if not impossible, to obtain supervisory signals and train deep detection models in a similar way as other tasks like classification and regression. 
While a wide range of unsupervised or self-supervised supervisory signals have been explored, as discussed in Section \ref{subsec:genericnormality}, to learn the representations, a key issue for these formulations is that their objective functions are generic but not optimized specifically for anomaly detection. Anomaly measure-dependent feature learning in Section \ref{subsec:measuredependent} helps address this issue by imposing constraints derived from traditional anomaly measures. However, these constraints can have some inherent limitations, \eg, implicit assumptions in the anomaly measures. It is critical to explore \textit{new sources of anomaly-supervisory signals} that lie beyond the widely-used formulations such as data reconstruction and GANs, and have weak assumptions on the anomaly distribution. Another possibility is to develop \textit{domain-driven anomaly detection} by leveraging domain knowledge \cite{dddm_Cao10} such as application-specific knowledge of anomaly and/or expert rules as the supervision source.

\subsection{Deep Weakly-supervised Anomaly Detection}

\textit{Deep weakly-supervised anomaly detection} \cite{pang2019ranking} aims at leveraging deep neural networks to learn anomaly-informed detection models with some weakly-supervised anomaly signals, \eg, partially/inexactly/inaccurately labeled anomaly data. This labeled data provides important knowledge of anomaly and can be a major driving force to lift detection recall rates \cite{pang2018repen,pang2019ranking,pang2019devnet,sultani2018ranking,tamersoy2014guilt}. One exciting opportunity is to utilize a small number of accurate labeled anomaly examples to enhance detection models as they are often available in real-world applications, \eg, some intrusions/frauds from deployed detection systems/end-users and verified by human experts. However, since anomalies can be highly heterogeneous, there can be unknown/novel anomalies that lie beyond the span set of the given anomaly examples. Thus, one important direction here is \textit{unknown anomaly detection}, in which we aim to build detection models that are generalized from the limited labeled anomalies to unknown anomalies. Some recent studies \cite{pang2019ranking,ruff2020deepsvdd,pang2019devnet,pang2020deep} show that deep detection models are able to learn abnormality that lie beyond the scope of the given anomaly examples. It would be important to further understand and explore the extent of the generalizability and to develop models to further improve the accuracy performance. 
% This is a major obstacle in this research line, leading to a significantly less explored direction compared to other directions.

To detect anomalies that belong to the same classes of the given anomaly examples can be as important as the detection of novel/unknown anomalies. Thus, another important direction is to develop \textit{data-efficient anomaly detection} or \textit{few-shot anomaly detection}, in which we aim at learning highly expressive representations of the known anomaly classes given only limited anomaly examples \cite{pang2018repen,pang2019ranking,tian2020fewshot,pang2019devnet}. It should be noted that the limited anomaly examples may come from different anomaly classes, and thus, exhibit completely different manifold/class features. This scenarios is fundamentally different from the general few-shot learning \cite{wang2020generalizing}, in which the limited examples are class-specific and assumed to share the same manifold/class structure. Additionally, as shown in Table \ref{tab:method}, the network architectures are mostly not as deep as that in other machine learning tasks. This may be partially due to the limitation of the labeled training data size. It is important to explore the possibility of leveraging those small labeled data to learn more powerful detection models with deeper architectures. Also, inexact or inaccurate (\eg, coarse-grained) anomaly labels are often inexpensive to collect in some applications \cite{sultani2018ranking}; learning deep detection models with this weak supervision is important in these scenarios.

\subsection{Large-scale Normality Learning}

% Most of deep semi-supervised anomaly detection assumes the given training data contains purely normal instances without any noise, so they are generally vulnerable to anomaly contamination in the training data. In real-world applications, it is hard to guarantee that a large-scale dataset is free of anomaly contamination. Thus, it is crucial to learn deep semi-supervised detection models that are robust to anomaly contamination. Also, when there are labeled anomaly examples available, it is important to incorporate those labeled data to learn anomaly-informed detection models \cite{ruff2020deepsvdd}. 

Large-scale unsupervised/self-supervised representation learning has gained tremendous success in enabling downstream learning tasks \cite{peters2018deep,devlin2018bert}. This is particular important for learning tasks, in which it is difficult to obtain sufficient labeled data, such as anomaly detection (see Section \ref{subsec:complexity}). The goal is to first learn transferable pre-trained representation models from large-scale unlabeled data in an unsupervised/self-supervised mode, and then fine-tune detection models in a semi-supervised mode. The self-supervised classification-based methods in Section \ref{subsec:selfprediction} may provide some initial sources of supervision for the normality learning. However, precautions must be taken to ensure that (i) the unlabeled data is free of anomaly contamination and/or (ii) the representation learning methods are robust w.r.t. possible anomaly contamination. This is because most methods in Sections \ref{sec:normality} implicitly assume that the training data is clean and does not contain any noise/anomaly instances. This robustness is important in both the pre-trained modeling and the fine-tuning stage. Additionally, anomalies and datasets in different domains vary significantly, so the large-scale normality learning may need to be domain/application-specific.

\subsection{Deep Detection of Complex Anomalies}

% Most existing deep anomaly detection methods are focused on point anomalies. However, there are other types of anomalies, \ie, conditional anomaly and group anomaly. These two types of anomalies exhibit completely different behaviors from the point anomalies, so existing methods are ineffective in detecting these anomalies. One main challenge here is to incorporate the concept of conditional/group anomalies into the representation or anomaly score learning. This may require very different approaches from the aforementioned approaches. For example, to detect group anomalies, we need to learn representations of a set of unordered data points rather than single data points. 

Most deep anomaly detection methods focus on point anomalies, showing substantially better performance than traditional methods. However, deep models for conditional/group anomalies have been significantly less explored. Deep learning has superior capability in capturing complex temporal/spatial dependence and learning representations of a set of unordered data points; it is important to explore whether deep learning could also gain similar success in detecting such complex anomalies. Novel neural network layers or objectives functions may be required.

Similar to traditional methods, current deep anomaly detection mainly focus on single data sources. \textit{Multimodal anomaly detection} is a largely unexplored research area. It is difficult for traditional approaches to bridge the gap presented by those multimodal data. Deep learning has demonstrated tremendous success in learning feature representations from different types of raw data for anomaly detection \cite{ding2019graphae,ionescu2019featureextraction,lu2017sequenceae,pang2018repen,sabokrou2018oneclasse2e}; it is also able to concatenate the representations from different data sources to learn unified representations \cite{goodfellow2016deep}, so deep approaches present important opportunities of multimodal anomaly detection.

\subsection{Interpretable and Actionable Deep Anomaly Detection}\label{subsec:interpretablemodel}

% Like other deep learning systems, in some critical domains there may be some major risks if deep detection models are directly used as black-box model. For example, since deep anomaly detection methods are optimized to identify \textit{rare} data instances as anomalies, one potential risk here is the possible algorithmic bias against the minority groups presented in the data, such as under-represented groups in fraud detection and crime detection systems. We may mitigate some parts of this risk by training the model with some known anomaly examples, which informs the model to detect the anomalies of our interest rather than simply rare instances. A more effective approach is to have anomaly explanation algorithms that provide straightforward clues about why a specific data instance is identified as anomaly. In important applications anomaly explanation is often as important as detection accuracy. 

Current deep anomaly detection mainly focuses on the detection accuracy aspect. \textit{Interpretable deep anomaly detection} and \textit{actionable deep anomaly detection} are essential for understanding model decisions and results, mitigating any potential bias/risk against human users and enabling decision-making actions. In recent years, there have been some studies \cite{angiulli2009explanation,duan2015explanation,vinh2016explanation,angiulli2017explanation,siddiqui2019explanation} that explore the anomaly explanation issues by searching for a subset of features that makes a reported anomaly most abnormal. The abnormal feature selection methods \cite{pang2016feature,azmandian2012feature,pang2017feature} may also be utilized for anomaly explanation purpose. The anomalous feature searching in these methods is independent from the anomaly detection methods, and thus, may be used to provide explanation of anomalies identified by any detection methods, including deep models. On the other hand, this model-agnostic approach may render the explanation less useful, because they cannot provide a genuine understanding of the mechanisms underlying specific detection models, resulting in weak interpretability and actionability (\eg, quantifying the impact of detected anomalies and mitigation actions).
% as our primary motivation is to understand why the instances are identified as anomalies by specific methods. 
% Many anomaly explanation studies \cite{angiulli2009explanation,duan2015explanation,vinh2016explanation,angiulli2017explanation,siddiqui2019explanation} pursue model-agnostic approaches, which may be used to provide explanation of anomalies identified by any anomaly detection methods, including deep models. On the other hand, they cannot provide a genuine understanding of the mechanisms underlying the detection models, especially for black-box models like deep learning-based models, resulting in potential modeling bias/risk and weak interpretability and actionability (\eg, quantifying the impact of detected anomalies and mitigation actions). 
Deep models with inherent capability to provide anomaly explanation is important, such as \cite{pang2020ranking}. To achieve this, methods for deep model explanation \cite{du2019interpretable} and actionable knowledge discovery \cite{dddm_Cao10} could be explored with deep anomaly detection models.

\subsection{Novel Applications and Settings}

There have been some exciting emerging research applications and problem settings, into which there could be some important opportunities of extending deep detection methods. First, \textit{out-of-distribution (OOD) detection} \cite{hendrycks2017ood,lee2018ood,ren2019ood} is a closely related area, which detects data instances that are drawn far away from the training distribution. This is an essential technique to enable machine learning systems to deal with instances of novel classes in open-world environments. OOD detection is also an anomaly detection task, but in OOD detection it is generally assumed that fine-grained normal class labels are available during training, and we need to retain the classification accuracy of these normal classes while performing accurate OOD detection. Second, \textit{curiosity learning} \cite{pathak2017curiosity,burda2019curiosity,burda2019curiosityb} aims at learning a bonus reward function in reinforcement learning with sparse rewards. Particularly, reinforcement learning algorithms often fail to work in an environment with very sparse rewards. Curiosity learning addresses this problem by augmenting the environment with a bonus reward in addition to the original sparse rewards from the environment. This bonus reward is defined typically based on the novelty or rarity of the states, \ie, the agent receives large bonus rewards if it discovers novel/rare states. The novel/rare states are concepts similar to anomalies. Therefore, it would be interesting to explore how deep anomaly detection could be utilized to enhance this challenging reinforcement learning problem; conversely, there can be opportunities to leverage curiosity learning techniques for anomaly detection, such as the method in \cite{wang2020distance}. Third, most shallow and deep models for anomaly detection assume that the abnormality of data instances is independent and identically distributed (IID), while the abnormality in real applications may suffer from some non-IID characteristics, \eg, the abnormality of different instances/features is interdependent and/or heterogeneous \cite{pang2019noniid}. For example, the abnormality of multiple synchronized disease symptoms is mutually reinforced in early detection of diseases. This requires \textit{non-IID anomaly detection} \cite{pang2019noniid} that is dedicated to learning such non-IID abnormality. This task is crucial in complex scenarios, \eg, where anomalies have only subtle deviations and are masked in the data space if not considering these non-IID abnormality characteristics. Lastly, other interesting applications include detection of adversarial examples \cite{grosse2017adversarial,paudice2018adversarial}, anti-spoofing in biometric systems \cite{perez2019spoofing,fatemifar2019spoofing}, and early detection of rare catastrophic events (\eg, financial crisis \cite{expert_CaoC15} and other black swan events \cite{aven2016risk}).
% by deep models and their combinations with shallow ones.

% \iffalse
% \section{Conclusions}

% We first discuss some unique problem complexities and largely unsolved challenges of anomaly detection. We then introduce a taxonomy of deep anomaly detection that groups the methods into three high-level categories and 11 fine-grained categories, representing 12 different perspectives on designing deep anomaly detection algorithms, and comprehensively review representative methods under this taxonomy and how they address the challenges. We further discuss the detailed implementation of these representative methods and their source codes and provide a list of real-world datasets with real anomalies to promote the development and evaluation of deep anomaly detection methods. We finally discuss five future research directions that present some exciting opportunities yet require more research efforts to tackle the relevant challenges.
% \fi

\appendix
\section{Links to Open-source Algorithms}\label{sec:sourcecode}

\begin{table}[htbp]
\centering
\caption{Links to Access 23 Open-source Algorithms.}
  \scalebox{0.65}{
    \begin{tabular}{cclc}
    \hline\hline
    \textbf{Method} & \textbf{API} & \textbf{Link} & \textbf{Section}\\\hline
    RDA \cite{zhou2017autoencoder} & Tensorflow & https://git.io/JfYG5 & \ref{subsubsec:ae}\\
    AnoGAN \cite{schlegl2017generative} & Tensorflow & https://git.io/JfGgc& \ref{subsubsec:generative}\\
    Fast AnoGAN \cite{schlegl2019generative} & Tensorflow & https://git.io/JfZRn & \ref{subsubsec:generative}\\
    EBGAN \cite{zenati2018generative} & Keras &  https://git.io/JfGgG & \ref{subsubsec:generative}\\
    ALAD \cite{zenati2018generativeicdm} & Keras & https://git.io/JfZ8v& \ref{subsubsec:generative}\\
    GANomaly \cite{akcay2018generative} & PyTorch & https://git.io/JfGgn & \ref{subsubsec:generative}\\
    FFP \cite{liu2018predictabilitycvpr} & Tensorflow & https://git.io/Jf4pc & \ref{subsec:selfprediction}\\
    LSA \cite{abati2019predictabilitycvpr} & Torch & https://git.io/Jf4pW & \ref{subsec:selfprediction}\\
    GT \cite{golan2018featuretrasnformation} & Keras & https://git.io/JfZRW & \ref{subsubsec:consistency}\\
    E$^3$Outlier \cite{wang2019featuretransformation} & PyTorch & https://git.io/Jf4pl  &  \ref{subsubsec:consistency}\\
    REPEN \cite{pang2018repen} & Keras & https://git.io/JfZRg & \ref{subsubsec:distancemeasure}\\
    RDP \cite{wang2020distance} & PyTorch & https://git.io/RDP & \ref{subsubsec:distancemeasure}\\
    AE-1SVM \cite{nguyen2018oneclasssvm} &Tensorflow & https://git.io/JfGgl & \ref{subsubsec:oneclass}\\
    OC-NN \cite{chalapathy2018oneclasssvm} & Keras & https://git.io/JfGgZ & \ref{subsubsec:oneclass}\\
    Deep SVDD \cite{ruff2018deepsvdd} & Tensorflow & https://git.io/JfZRR & \ref{subsubsec:oneclass}\\
    Deep SAD \cite{ruff2020deepsvdd} & PyTorch & https://git.io/JfOkr  & \ref{subsubsec:oneclass}\\
    DAGMM \cite{zong2018autoencoder} & PyTorch &  https://git.io/JfZR0 & \ref{subsubsec:clustering}\\
    MIL \cite{sultani2018ranking} & Keras & https://git.io/JfZRz & \ref{subsec:ranking}\\
    DevNet \cite{pang2019devnet} & Keras & https://git.io/JfZRw & \ref{subsec:prior} \\
    ALOCC \cite{sabokrou2018oneclasse2e} & Tensorflow & https://git.io/Jf4p4 & \ref{subsec:oneclasse2e}\\
    OCAN \cite{zheng2019oneclasse2e} & Tensorflow & https://git.io/JfYGb & \ref{subsec:oneclasse2e}\\
    FenceGAN \cite{ngo2019oneclasse2e} & Keras & https://git.io/Jf4pR & \ref{subsec:oneclasse2e}\\
    OCGAN \cite{perera2019oneclasse2e} & MXNet  & https://git.io/Jf4p0 & \ref{subsec:oneclasse2e}\\
    \hline\hline
    \end{tabular}
    }
\label{tab:code}
\end{table}

\bibliographystyle{ACM-Reference-Format}
\bibliography{reference}

\end{document}